\documentclass{article}

\usepackage{arxiv}

\usepackage[utf8]{inputenc} 
\usepackage[T1]{fontenc}    
\usepackage{hyperref}       
\usepackage{url}            
\usepackage{booktabs}       
\usepackage{amsfonts}       
\usepackage{amsmath,bm,upgreek}
\usepackage{amsthm}
\usepackage{nicefrac}       
\usepackage{microtype}      
\usepackage{lipsum}
\usepackage{graphicx}
\graphicspath{ {./images/} }
\usepackage{algorithm}%
\usepackage{algorithmicx}
\usepackage{algpseudocode}%
\usepackage{xcolor}
\usepackage{subcaption}

\def \by {\mathbf{y}}

\def \bx {\mathbf{x}}

\def \bu {\mathbf{u}}

\def \bk {\mathbf{k}}

\def \calL {\mathcal{L}}

\def \Cov {\mathrm{Cov}}

\newtheorem{theorem}{Theorem}

\title{Physics-informed Gaussian Process Regression in Solving Eigenvalue Problem of Linear Operators}

\author{
 Tianming Bai \\
  School of AI and Advanced Computing\\
  Xi'an Jiaotong Liverpool University\\
  Suzhou, China\\
  \texttt{Tianming.Bai@xjtlu.edu.cn} \\
   \And
 Jiannan Yang \\
  School of Physics, Engineering and Technology\\
  University of York\\
  York, UK\\
  \texttt{jiannan.yang@york.ac.uk} \\
}

\begin{document}
\maketitle
\begin{abstract}
Applying Physics-Informed Gaussian Process Regression to the eigenvalue problem $(\mathcal{L}-\lambda)u = 0$ poses a fundamental challenge, where the null source term results in a trivial predictive mean and a degenerate marginal likelihood. Drawing inspiration from system identification,  we construct a transfer function-type indicator for the unknown eigenvalue/eigenfunction using the physics-informed Gaussian Process posterior. We demonstrate that the posterior covariance is only non-trivial when $\lambda$ corresponds to an eigenvalue of the partial differential operator $\mathcal{L}$, reflecting the existence of a non-trivial eigenspace, and any sample from the posterior lies in the eigenspace of the linear operator. We demonstrate the effectiveness of the proposed approach through several numerical examples with both linear and non-linear eigenvalue problems. 
\end{abstract}


\section{Introduction}
Gaussian Process Regression (GPR) is a powerful non-parametric Bayesian method that provides a flexible framework for function approximation \cite{rasmussen2003gaussian}. A key advantage of GPR is its inherent ability to quantify uncertainty, offering a full probabilistic posterior distribution over the unknown function. In recent years, this framework has been extended to incorporate physical laws, leading to Physics-Informed Gaussian Process Regression (PI-GPR) \cite{RRPK17,cockayne2016probabilistic}. PI-GPR leverages the property that bounded linear operators, such as differential operators, applied to a GP result in another GP \cite{PSH22}. This allows for the direct encoding of a Partial Differential Equation's (PDE) structure into the covariance function, conditioning the prior on the governing physics.

Eigenvalue problems governed by partial differential operators, expressed as $\mathcal{L}u = \lambda u$, are fundamental in virtually all areas of physics and engineering. They are essential for modelling critical phenomena such as natural frequencies in vibration analysis \cite{inman1994engineering}, energy levels in quantum mechanics \cite{griffiths2018introduction}, and buckling loads in structural stability \cite{timoshenko2012theory}. Solving these problems requires finding the eigenvalues $\lambda$, for which the system admits non-trivial solutions, the eigenfunctions $u$. The pursuit of new computational methods for these problems remains an active area of research, particularly methods that can integrate data or provide inherent measures of uncertainty.

Despite the success of PI-GPR for standard boundary value problems (BVPs) with non-zero source terms, its application to the eigenvalue problem presents a fundamental challenge. For example, when a natural zero-mean GP prior is placed on the eigenfunction $u$, the posterior mean becomes trivially zero, regardless of the value of $\lambda$. This collapse renders the predictive mean uninformative. Meanwhile, it also nullifies the data-fit term in the negative log-marginal likelihood (NLML), the standard objective function for hyperparameter optimisation and model selection (see for instance Eq.\ref{eq:nlml}). Consequently, the NLML cannot be used to infer the eigenvalue $\lambda$.

The challenge of combining probabilistic methods with eigenvalue problems has been approached from several perspectives. For instance, in the stochastic eigenvalue problem (SEP), the focus is on how uncertainty in the operator $\mathcal{L}$ or the domain $\Omega$ propagates to the eigenvalues $\lambda$ \cite{ghanem2003stochastic}. Other machine learning approaches have focused on creating emulators for parametric eigenvalue problems. For example, Bertrand et al. \cite{bertrand2023data} proposed a non-intrusive reduced basis method using Proper Orthogonal Decomposition (POD) snapshots and GPR. Their GPR model learns to predict the modal coefficients for a POD basis, but its predictions are purely data-driven and constrained by the span of the initial snapshots. These approaches differ from our goal, which is to use the PI-GPR framework to discover the eigenvalues of a single, deterministic operator from its physical definition alone.

In this paper, we propose a new probabilistic method to solve the eigenvalue problem of linear PDEs by constructing a transfer function-type indicator for the unknown eigenvalue/eigenfunctions using the PI-GPR posterior. We demonstrate that the posterior covariance is only non-trivial when $\lambda$ corresponds to an eigenvalue of the PDE, and any sample from the posterior lies in the eigenspace of the linear
operator. That is our main contribution. 

More specifically, we demonstrate that for the eigenvalue problem, the key information lies not in the posterior mean, but in the posterior covariance. Our central thesis is that the "size" of the posterior covariance serves as a powerful indicator for eigenvalues. Heuristically, when the parameter $\lambda$ is not an eigenvalue, the operator $(\mathcal{L}-\lambda)$ admits only the trivial solution $u=0$. The PI-GPR framework correctly identifies this, and the posterior covariance is close to zero. Conversely, when $\lambda$ is an eigenvalue, the operator admits a non-trivial null space (the eigenspace). The GP posterior, conditioned on the physics, becomes a non-trivial distribution over this eigenspace, and its covariance becomes non-zero. 

To substantiate this heuristic, we first provide a rigorous proof for the finite-dimensional analogue where the operator is a matrix. Leveraging the properties of the conditional Gaussian distribution \cite{eaton1983multivariate} and the Moore-Penrose pseudoinverse \cite{moore1920reciprocal, penrose1955generalized}, Theorem \ref{thm:samp} establishes the analytical link between the operator singularity and the posterior variance. While a complete operator-theoretic proof for the infinite-dimensional PI-GPR setting is beyond the scope of this work, the necessary theoretical foundation is established by Henderson et al \cite{henderson2023characterization}. 

We then develop a computational method based on this principle, where we scan a range of $\lambda$ values and compute the trace of the posterior covariance matrix $J(\lambda)$, which is a function of the unknown $\lambda$. The eigenvalues are then identified from the distinctive peaks of the spectrum of $J(\lambda)$. This approach draws inspiration from system identification, where the PDE is used as a filter for the GP prior so that the eigenfunctions can be identified. It is thus easy to implement and we provide several examples to demonstrate its effectiveness. 

The eigenvalue/eigenvectors of a PDE are a rich and significant field of study in their own right, providing insights into the nature of many physical problems. It should be noted that our method can also be seen as a precursor step to use PI-GPR to solve the homogeneous PDEs $Lu=0$, as the PDE's Green's function can often be expressed as an expansion of the associated eigenfunctions. Previous work has used PI-GPR to solve homogeneous PDEs $Lu=0$, often by constraining the covariance function using the homogeneous constraints. One approach is to make use of the fundamental solution or Green's function of the linear PDE. For example, constrained covariance functions are built for classical PDEs such as the Laplace equation and the Helmholtz equation \cite{albert2020gaussian}, and the detailed mathematical foundation of this approach is discussed in \cite{cockayne2016probabilistic}. Another approach uses the Ehrenpreis-Palamodov principle, where the PDE solutions are expressed as a superposition of exponential-polynomial functions, to construct the GP covariance functions \cite{harkonen2023gaussian}. The Ehrenpreis-Palamodov GP (EPGP) framework guarantees that GP solutions are members of the PDE null space, ensuring all samples inherently satisfy the governing physics. However, both approaches have computational limitations, as convolutions or multidimensional integrals are needed to encode the physics into the covariance function. In contrast, the proposed method in this paper is much more efficient as we can use generic covariance functions in a standard PI-GPR framework and there is no convolution involved. 

The remainder of this paper is organised as follows. Section \ref{sec:2} provides preliminaries on GPR and the PI-GPR framework. Section \ref{sec:met} details our covariance-based methodology, presenting the theoretical motivation from a finite-dimensional perspective and the algorithm for the infinite-dimensional PDE case. Section \ref{sec:num} presents numerical experiments to validate our approach, demonstrating its ability to accurately identify eigenvalues for several canonical problems. Finally, Section \ref{sec:conclusion} offers concluding remarks and directions for future research.

\section{Preliminaries}
\label{sec:2}
\subsection{Gaussian Process Regression}
GPR is a powerful Bayesian non-parametric method widely used for function approximation, uncertainty quantification, and inverse problems. Given a set of observational data, GPR assumes that the underlying unknown function is a realisation of a GP, fully characterised by its mean function $m(\bx)$ and covariance function $k(\bx,\bx')$. Formally, we write
\begin{equation}
u(\bx) \sim \textrm{GP}(m(\bx), k(\bx,\bx')).
\end{equation}
A common choice for the mean function $m(\bx)$ is a zero function, reflecting no prior knowledge. For covariance functions, two frequently used choices are the squared exponential (SE) covariance function and the Mat\'ern covariance function. The squared exponential covariance function is defined as
\begin{equation}
k_{SE}(\bx, \bx') = \sigma^2 \exp\left(-\frac{|\bx - \bx'|^2}{2\ell^2}\right),
\end{equation}
where $\sigma^2$ controls the variance and $\ell$ is the length-scale. While other covariance functions such as the Matérn class \cite{matern2013spatial, rasmussen2003gaussian}, offer flexibility in smoothness, in this paper we will focus on the SE covariance function for numerical demonstrations with canonical physics problems. In addition, the use of SE covariance functions allows us to investigate the spectral properties of the eigenvalue solutions as shown in Section \ref{sec:num}.


Given noise-free observational data $u(X)$ at spatial points $u(X) = \{u(\bx_i)\}_{i=1}^{N} \in \mathbb{R}^N$, the predictive distribution for the function $u$ at a new point $\bx$ is a GP with posterior mean and covariance given by:
\begin{align}
m_{N}(\bx) &= m(\bx) + k(\bx,X)k(X,X)^{-1}(u(X)-m(X)),\\
k_{N}(\bx, \bx') &= k(\bx,\bx') - k(\bx,X)k(X,X)^{-1}k(X,\bx').
\end{align}
In this case, we have full confidence in the observed data, resulting in zero uncertainty at the observed points and uncertainty appearing only in unobserved regions. When observations are noisy, the observational model is typically written as $\mathbf{y} = u(X) + \xi$, where $\xi\sim \mathcal{N}(0,\Sigma_\xi)$. The predictive distribution in this noisy scenario becomes:
\begin{align}
m_{N}(\bx) &= m(\bx) + k(\bx,X)\left(k(X,X) + \Sigma_\xi\right)^{-1}(\mathbf{y}-m(X)),\\
k_{N}(\bx, \bx') &= k(\bx,\bx') - k(\bx,X)\left(k(X,X) + \Sigma_\xi\right)^{-1}k(X,\bx').
\end{align}

\subsection{Physics-Informed Gaussian Process Regression}
When using GPR to solve Partial Differential Equations (PDEs), one can exploit the property that the linear transformation of a GP with standard covariance functions (e.g., squared exponential or Mat\'ern covariance functions) yields another GP \cite{PSH22}. This property allows for constructing joint and conditional GPs relating different quantities within linear differential equations. This approach is known as PI-GPR. Specifically, consider the PDE given by
\begin{equation}\label{eq:pde_general}
\calL u(\bx) = f(\bx),
\end{equation}
where $\calL$ is a known linear differential operator, $u$ is the unknown solution, and $f$ is a known source term. In PI-GPR, we first assume the GP prior on $u$ as
$u(\bx) \sim \textrm{GP}(m(\bx), k(\bx,\bx'))$.
Utilising the property, we obtain the prior on the source term as
\begin{equation}\label{eq:prior_rhs}
f(\bx) \sim \textrm{GP}(\calL m(\bx), \calL \calL 'k(\bx,\bx')),
\end{equation}
where the prime symbol ``$\, '\,$'' denotes the operator applied to the second argument $\bx'$ of $k(\bx,\bx')$. Given point evaluations of the source term $f$, specifically, $\mathbf{f}(X_f) \in \mathbb{R}^{N_f}$ where $X_f = \{\bx_i\}_{i=1}^{N_f}$, we obtain the conditional GP
\begin{equation}\label{eq:PI_Gpr}
u_{N_f}(\bx) = u(\bx) | \mathbf{f}(X_f) \sim \textrm{GP}(m_{N_f}(\bx), k_{N_f}(\bx,\bx')),
\end{equation}
with conditional mean and covariance functions:
\begin{equation}\label{eq:post_mean}
m_{N_f}(\bx) = m(\bx) + \calL '\bk(\bx, X_f)\left(\calL \calL 'K(X_f,X_f)\right)^+\left(\mathbf{f}(X_f) - \calL m(X_f)\right),
\end{equation}
and
\begin{equation*}\label{eq:post_cov}
k_{N_f}(\bx, \bx') = k(\bx, \bx') - \calL '\bk(\bx, X_f)\left(\calL \calL 'K(X_f,X_f)\right)^+\calL \bk(X_f,\bx'),
\end{equation*}
where the superscript ``$^+$'' denotes the Moore-Penrose pseudo-inverse \cite{eaton1983multivariate,moore1920reciprocal}. Since applying bounded linear operators on covariance functions preserves only positive semi-definiteness, the Moore-Penrose pseudo-inverse is used for the possibility that \(\calL \mathcal{L'}K(X_{f}, X_{f})\) is singular \cite{scholkopf2002learning}. The vectors \(\mathcal{L'}\bk(\bx,X_{f})^{\top}\) and \(\calL \bk(X_{f},\bx)\) of size \(N_f\) have respective \(i_{th}\) entries \(\mathcal{L'}k(\bx,\bx_i)\) and \(\calL k(\bx_i,\bx)\) with $\bx_i \in X_f$. The matrix \(\calL \mathcal{L'}K(X_{f},X_{f})\) of size \(N_f\times N_f\) has its \((i,j)_{th}\) entry \(\calL \mathcal{L'}k(\bx_i, \bx_j)\) with $\bx_i, \bx_j \in X_f$. 

Similarly, boundary conditions can be incorporated seamlessly in the PI-GPR framework. If the boundary conditions are linear (which is often the case, such as Dirichlet or Neumann conditions), they can be expressed as $\mathcal{B}u(\mathbf{x}) = g(\mathbf{x})$, where $\mathcal{B}$ is a linear operator and $g$ is a known function. Utilising the linear transformation property of GPs, the boundary constraints can be similarly imposed to refine the posterior. Crucially, the information from the PDE (via $\mathcal{L}$) and the information from the boundary conditions (via $\mathcal{B}$) can be embedded simultaneously within the joint GP framework, allowing for a comprehensive, physics-constrained approximation of the solution $u$.

\subsection{Toy example}
We now illustrate the PI-GPR method with a simple numerical experiment. Consider a 1-dimensional differential equation with Dirichlet boundary conditions
\begin{equation*}
\begin{aligned}
&-\frac{\mathrm{d}^2}{\mathrm{d}x^2}u(x) = 10, \quad x \in (0,1), \\
&\quad u(0) = 0, \quad  u(1) = 0. 
\end{aligned}   
\end{equation*}
The exact solution of this boundary value problem is
\begin{equation}\label{eq:true_sol}
 u(x) = - 5x^2 + 5x. 
\end{equation}

\begin{figure}[htbp]
    \centering    
    \begin{flushright}
    \includegraphics[height=2.0cm]{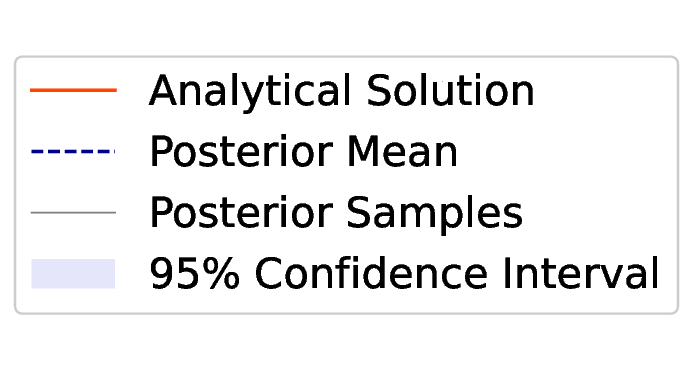} 
    \end{flushright}
    \vspace{-0.5cm} 
    \begin{subfigure}[b]{0.4\textwidth}
        \centering        \includegraphics[width=\textwidth]{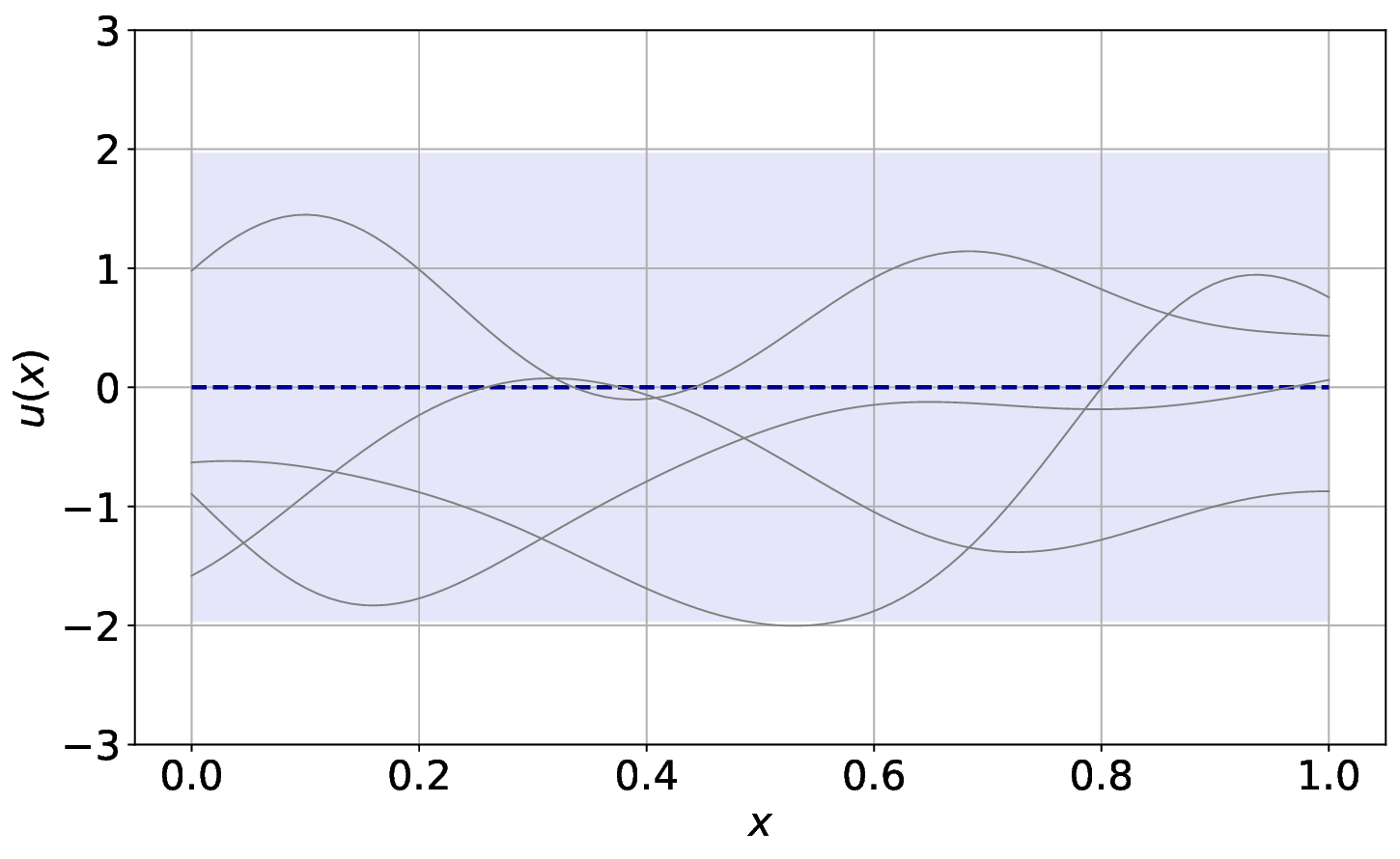}   \renewcommand\thesubfigure{a}
        \caption{}
        \label{fig:pre_a}
    \end{subfigure}
    \begin{subfigure}[b]{0.4\textwidth}
        \centering
        \includegraphics[width=\textwidth]{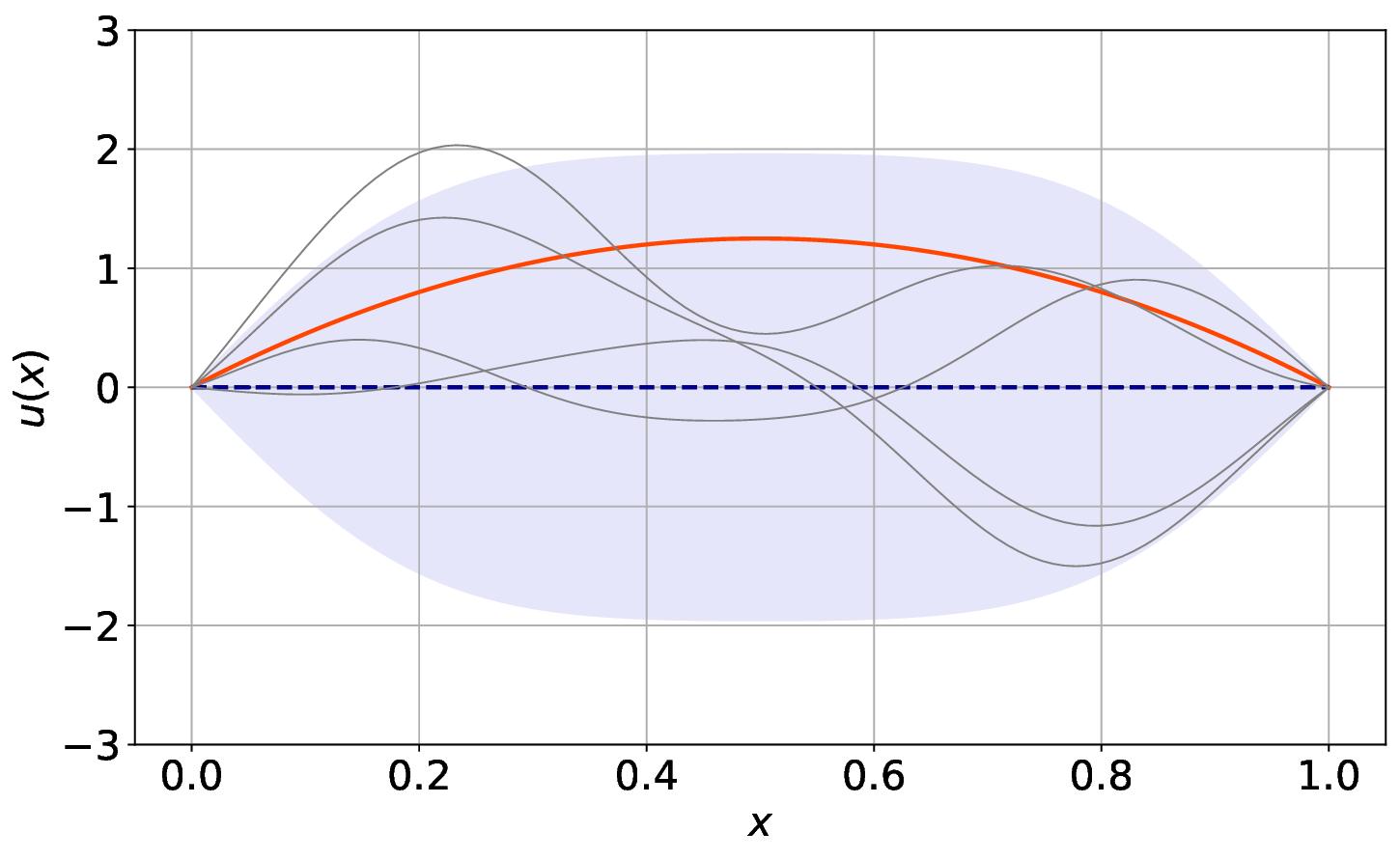}   \renewcommand\thesubfigure{b}
        \caption{}
        \label{fig:pre_b}
    \end{subfigure}    
    \begin{subfigure}[b]{0.4\textwidth}
        \centering        \includegraphics[width=\textwidth]{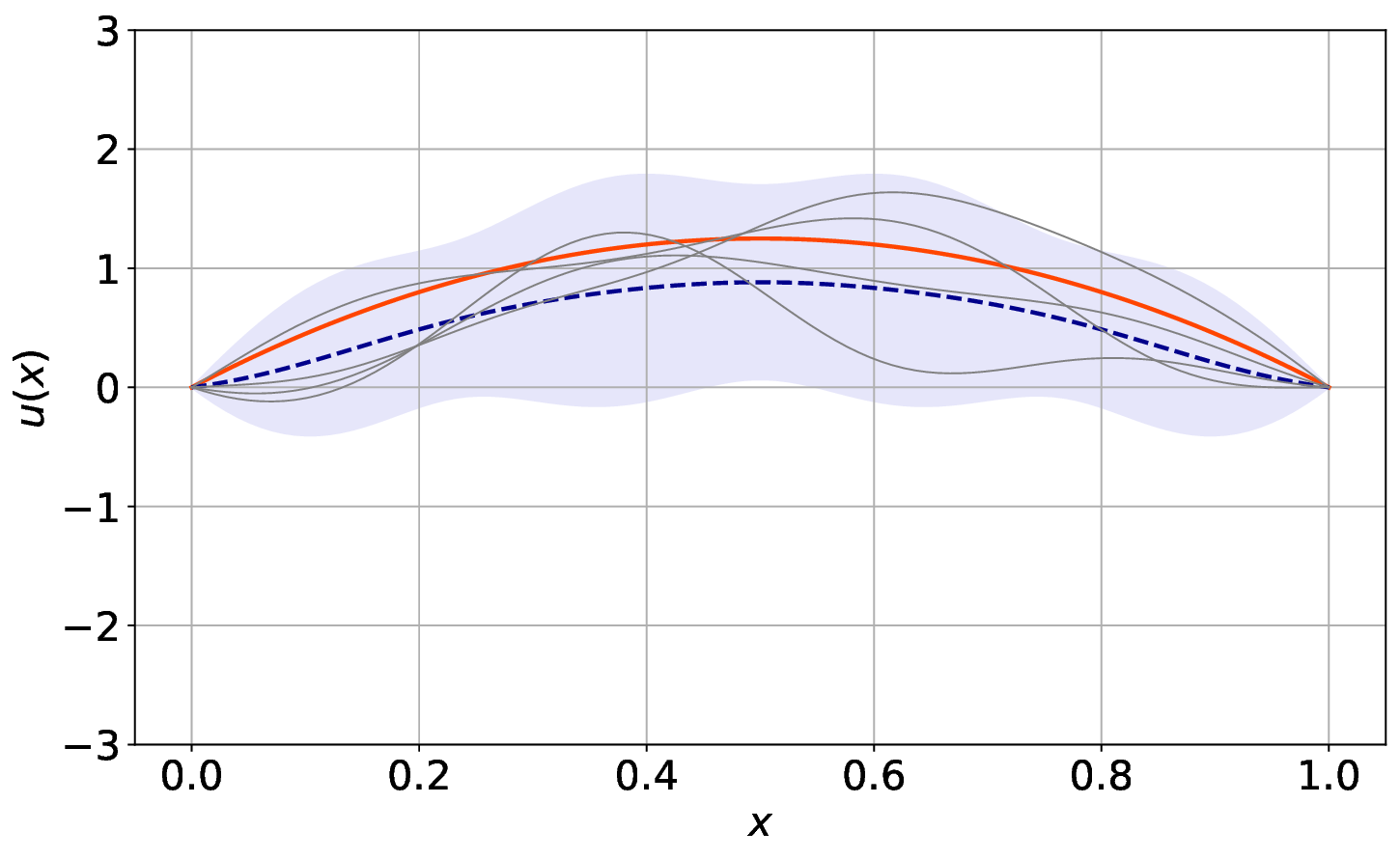}   \renewcommand\thesubfigure{c}
        \caption{}
        \label{fig:pre_c}
    \end{subfigure}
    \begin{subfigure}[b]{0.4\textwidth}
        \centering        \includegraphics[width=\textwidth]{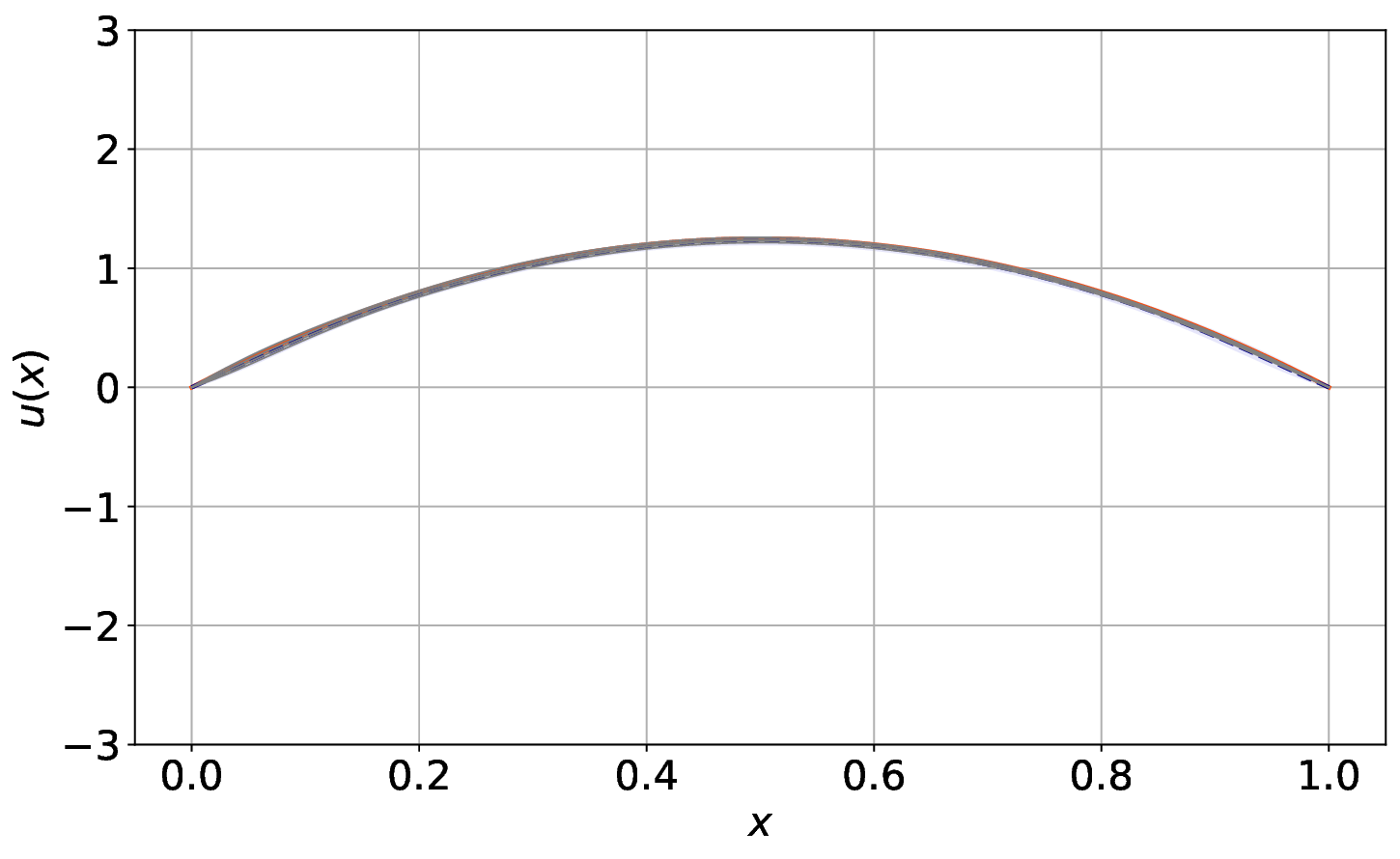}   \renewcommand\thesubfigure{d}
        \caption{}
        \label{fig:pre_d}
    \end{subfigure}
\caption{The mean and samples of the GPs: (a) Prior; (b) Conditioned on two boundary points; (c) Conditioned on boundary points and $N_f = 3$ collocation points from $f$; (d) Conditioned on two boundary points and $N_f = 8$ from $f$. The grey lines are samples from the posterior, the dark blue dashed lines represent the posterior mean, the orange lines indicate the exact solution Eq.\eqref{eq:true_sol}, and the purple regions represent the $95\%$ confidence interval.}
\label{fig:PDE_prior_samples}
\end{figure}

To approximate this solution via PI-GPR, we first assign a GP prior to $u$ as $u(\bx) \sim \textrm{GP}(0, k(\bx,\bx'))$, utilising a zero mean and a squared exponential covariance function. The hyperparameters are set to $\sigma = 1$ and length-scale $l = 0.2$. Samples drawn from this unconstrained GP prior are shown in Figure \ref{fig:PDE_prior_samples} (top left). Next, we constrain the prior by enforcing the boundary conditions. Conditioning on these boundary points collapses the uncertainty to zero at the domain endpoints, forcing all samples to intersect these fixed points (top right).

We subsequently constrain the prior by incorporating PDE information at $X_f$, representing equally spaced points within $[0,1]$ (excluding boundaries). As we increase $N_f$ from $0$ (top right) to $3$ (bottom left), the predictive mean converges noticeably toward the true solution. At $N_f = 8$ (bottom right), both the samples and the mean become virtually indistinguishable from the exact solution.

It can be seen from this example that standard application of PI-GPR typically requires a known non-trivial source term, such as the constant curvature in this example, to constrain the GP prior and update the posterior mean. Its direct application to the eigenvalue problem presents a fundamental challenge, as both sides of the eigenvalue equation explicitly depend on the unknown solution.

\section{Methodology}
\label{sec:met}
\subsection{Problem Formulation within the Physics-Informed GPR Framework}

The canonical form of a PDE eigenvalue problem is to find the eigenvalues $\lambda$ and corresponding non-trivial eigenfunctions $u(\mathbf{x})$ that satisfy:
\begin{equation}\label{eq:eig_original}
\begin{aligned}
    \calL  u(\mathbf{x}) &= \lambda u(\mathbf{x}), & \qquad \mathbf{x} \in \Omega \\
    \mathcal{B}u(\mathbf{x}) &= 0, & \qquad \mathbf{x} \in \partial\Omega
\end{aligned}
\end{equation}
where $\calL$ is a linear differential operator and $\mathcal{B}$ defines the homogeneous boundary conditions on the domain $\Omega$. The system is posed on the domain $\Omega$. A key distinction from standard boundary value problems (BVPs) is the pursuit of non-trivial solutions ($u \neq 0$), which typically requires a normalisation constraint (e.g., $\|u\|_{L^2}=1$) for uniqueness.

In applying PI-GPR to solve BVPs, we typically use observational data derived from a known source term $f$ in Eq.\eqref{eq:pde_general}. This data provides physics information for the regression, leading to meaningful predictions. However, in the eigenvalue problem Eq.\eqref{eq:eig_original}, both sides explicitly depend on the unknown solution $u$. To apply the PI-GPR framework, we reformulate the eigenvalue problem into a homogeneous form:
\begin{equation}\label{eq:eig_homogeneous}
(\calL - \lambda)u(\bx) = 0,
\end{equation}
and treat $\lambda$ as an unknown parameter to be inferred, analogous to an inverse problem. This reformulation deviates from the conventional PI-GPR framework and introduces three interconnected challenges:

First, a trivial GP posterior mean function. It is natural to place a zero-mean GP prior, $u \sim \mathcal{GP}(0, k(\mathbf{x}, \mathbf{x}'))$, on the eigenfunction, as any non-zero mean would impose a bias on a function defined only up to scale and sign. However, when both the prior mean and the observation data are zero, the posterior mean of the GP is also trivially zero. The standard predictive mechanism of GPR thus becomes uninformative, failing to provide the sought-after non-trivial eigenfunction.

Second, a new interpretation of the posterior covariance function is required. In standard GPR, the covariance quantifies the uncertainty of the prediction, which typically decreases as more data is acquired. Here, its role must be fundamentally different. Since the posterior mean is zero, the covariance becomes the sole carrier of information about the structure of the solution space. It must somehow characterise the existence of a non-trivial null space for the operator $(\calL  - \lambda I)$ when $\lambda$ is an eigenvalue.

Finally, the homogeneous nature of the problem invalidates the standard method for estimating the hyperparameters by minimising the negative log-marginal likelihood (NLML). The objective function of NLML is given by
\begin{equation}\label{eq:nlml}
    -\log p(\by|X) = \frac{1}{2}\by^{\top}K^{-1}\by + \frac{1}{2}\log |K| + \frac{n}{2}\log 2\pi.
\end{equation}
where $\by$ denotes the observational data, and $K_{\lambda}$ denotes its corresponding prior covariance matrix. This objective function includes a trade-off between the fitting of the observational data (the first term) and the complexity of the model (the second term). However, with observation data $\mathbf{y} = \mathbf{0}$, the crucial data-fit term vanishes. The objective function is no longer suitable for optimising hyperparameters and identifying the specific values of $\lambda$ that admit a non-trivial solution.

Given these observations, it can be seen that the trivial posterior GP mean compels a fundamental reinterpretation of the posterior covariance and that has motivated our research. In standard PI-GPR applications involving non-homogeneous PDEs, the operator is typically invertible and the solution is unique. Consequently, as the number of collocation points increases, the GP constraints progressively converge to this unique solution, causing the posterior variance to vanish. In this scenario, the posterior covariance represents the remaining uncertainty in the system. 

In contrast, the eigenvalue problem $(\mathcal{L}-\lambda)u = 0$ is fundamentally a search for singularities. When $\lambda$ is not an eigenvalue, the operator is invertible and the problem is well-posed with a unique trivial solution $u \equiv 0$. Here, the posterior correctly collapses to zero variance, reflecting certainty in the trivial outcome. However, when $\lambda$ is an eigenvalue, the operator becomes singular and the null space contains infinitely many solutions. Since the physical constraints do not enforce a unique amplitude, the posterior distribution must reflect the uncertainty in the inference of the eigenfunctions. Therefore, it is expected that a high posterior uncertainty is not a sign of poor approximation in the conventional GP sense, but the indication that the operator has admitted a non-trivial solution structure. We will make use of this unique structure and present a method to identify eigenvalues and eigenfunctions in the next section. 

\subsection{An Eigenvalue Criterion from the GP Posterior Covariance}

Our research reveals that the key to a successful application of PI-GPR to solve eigenvalue problems lies in examining the posterior covariance. Heuristically, the posterior covariance matrix, in a suitable norm, should be near-zero when $\lambda$ is not an eigenvalue, and non-zero (and large) when $\lambda$ is an eigenvalue. To examine this observation, we first present theoretical results obtained in the finite-dimensional setting of linear equation systems, and then discuss the infinite-dimensional PDE case with numerical examples.

Consider a linear system \(A\bu = \mathbf{f}\), where \(\bu \in \mathbb{R}^N\) and \(A\in \mathbb{R}^{m\times N}\). We place a multivariate Gaussian prior on the unknown vector  \(\bu \sim \mathcal{N}\left(0, K\right)\), where \(K\) is a positive definite covariance matrix. By the linear transformation of Gaussian random variables, we have the covariance \(\Cov(\bu,\mathbf{f}) = KA^{\top}\), \(\Cov(\mathbf{f},\bu) = AK\) and \(\Cov(\mathbf{f},\mathbf{f}) = AKA^{\top}\), so the joint distribution between \(\bu\) and \(\mathbf{f}\) is
\begin{equation*}
    \begin{bmatrix}
    \bu \\ \mathbf{f}
    \end{bmatrix}\sim \mathcal{N}\left(\boldsymbol{0}, \begin{bmatrix}
    K & KA^{\top} \\
AK & AKA^{\top}\\
    \end{bmatrix}\right).
\end{equation*}
Then we can compute the conditional distribution as 
\[\bu|\mathbf{f} \sim \mathcal{N}\left(KA^\top(AKA^\top)^{+}\mathbf{f}, K - KA^\top(AKA^\top)^{+}AK \right),\] where $(\cdot)^{+}$ represents the Moore-Penrose pseudo-inverse.

Our eigenvalue problem, $(L-\lambda I)\mathbf{u} = \mathbf{0}$, is a special case where $A = L-\lambda I$ and the conditioning data is $\mathbf{f}=\mathbf{0}$. Consequently, the posterior mean is trivially zero, and the posterior distribution is entirely characterised by its covariance:
\begin{equation}\label{equ:LA_post_cov}
\mathbf{u} | ((L-\lambda I)\mathbf{u} = \mathbf{0}) \sim \mathcal{N}\left(\mathbf{0}, K_{N}(\lambda) \right),
\end{equation}
where  $K_{N}(\lambda) = K - K(L-\lambda I)^\top((L-\lambda I)K(L-\lambda I)^\top)^{+}(L-\lambda I)K$. The behaviour of this posterior covariance $K_{N}(\lambda)$ changes dramatically depending on whether $\lambda$ is an eigenvalue of $L$. This is the foundation of our criterion, as summarised in the following theorem.

\begin{theorem}\label{thm:samp}
Let $A(\lambda) = L-\lambda I$. The posterior covariance matrix $$K_{N}(\lambda) = K - KA(\lambda)^\top(A(\lambda)KA(\lambda)^\top)^{+}A(\lambda)K$$ has the following properties (A formal proof is provided in Appendix \ref{appendix:theorem}):
\begin{itemize}
    \item If $\lambda$ \textbf{is not an eigenvalue} of $L$, the matrix $A(\lambda)$ has a trivial null space (is full rank). In this case, the posterior covariance $K_{N}(\lambda) = \mathbf{0}$. The posterior distribution collapses to the trivial solution $\mathbf{u}=\mathbf{0}$.
    \item If $\lambda$ \textbf{is an eigenvalue} of $L$, the matrix $A(\lambda)$ is rank-deficient and has a non-trivial null space, $\text{Null}(A(\lambda))$, which is the eigenspace. The posterior covariance $K_{N}(\lambda)$ is a non-zero matrix. Any vector sampled from the posterior distribution $\mathcal{N}(0,K_{N}(\lambda))$ is in the eigenspace of $L$ corresponding to the eigenvalue $\lambda$.
\end{itemize}
\end{theorem} 

This theorem reveals an indicator for eigenvalues. A scalar metric of the "size" of the posterior covariance, such as its trace, $J(\lambda) = \text{Tr}(K_{N}(\lambda))$, will be zero or near-zero when $\lambda$ is not an eigenvalue, but will become large in amplitude when $\lambda$ corresponds to an eigenvalue, thus providing an indicator for the eigenvalues. 

It is important to note that while the eigenspace itself is an unbounded subspace, the posterior does not assign uniform probability to it. The GP prior $\mathbf{u} \sim \mathcal{N}(0, K)$ defines a Gaussian probability measure over a space of candidate functions, where the covariance kernel $K$ implicitly provides a rich set of basis functions. Conditioning this measure on the homogeneous constraint $(\mathcal{L} - \lambda)\mathbf{u} = 0$ effectively acts as a functional filter, restricting the posterior to the null space of the operator. From this perspective, a non-vanishing posterior trace $J(\lambda)$ signifies that the prior measure has a non-trivial projection onto the eigenspace at $\lambda$, allowing the posterior to capture the inherent uncertainty in the amplitude of the eigenfunction.

The extension of our finite-dimensional analogy in Theorem \ref{thm:samp} to the infinite-dimensional setting of PI-GPR requires a rigorous functional analysis framework. While a complete operator-theoretic proof is beyond the scope of this paper, we present numerical illustrations in Section \ref{sec:num} using infinite-dimensional PDEs. In addition, recent work by Henderson et al \cite{henderson2023characterization} provides the necessary theoretical foundation. They analyse random fields whose sample paths satisfy a homogeneous PDE, $L(u)=0$, by interpreting the equation in the sense of distributions. This powerful functional analysis approach is crucial as it minimises the differentiability assumptions required for the random field's sample paths and moments. Their key result establishes a formal equivalence: the sample paths of the process (our eigenfunction $u$) satisfy the PDE almost surely if and only if its covariance function also satisfies the PDE in the same distributional sense. This theorem provides a formal justification for our method's core premise: we can infer properties of the solution $u$ by analysing the structure of the covariance.

Based on the trace criterion $J(\lambda)$, we can identify the eigenvalues of the system by performing a spectral scan across a candidate range of $\lambda$. In practice, the domain can be discretised using $N$ collocation points, and the posterior covariance is evaluated at $N_t$ test points to compute the metric $J(\lambda)$. The formal pseudo-code (Algorithm \ref{alg:eigen_scan_v2}) and a discussion on its computational complexity are provided in Appendix \ref{appendix:computation}.

\section{Numerical illustrations}\label{sec:num}
\subsection{One-dimensional Laplace's equation with Dirichlet boundary conditions}
\label{subsec:2nd order}
In this numerical experiment, we consider the one-dimensional eigenvalue problem associated with the negative Laplacian operator on the interval $[0, 1]$, subject to Dirichlet boundary conditions, that is
\begin{equation}\label{equ:lap}
    -\frac{d^2u}{dx^2} = \lambda u, \quad x \in (0, 1), \qquad u(0) = u(1) = 0.
\end{equation}
This problem admits an infinite sequence of positive eigenvalues and corresponding eigenfunctions given analytically by
\begin{equation}
    \lambda_n = \left( n\pi \right)^2, \qquad u_n(x) = \sin\left( n\pi x \right), \quad n = 1, 2, 3, \dots
\end{equation}

\begin{figure}[htbp]
    \centering
    \includegraphics[width=0.6\textwidth]{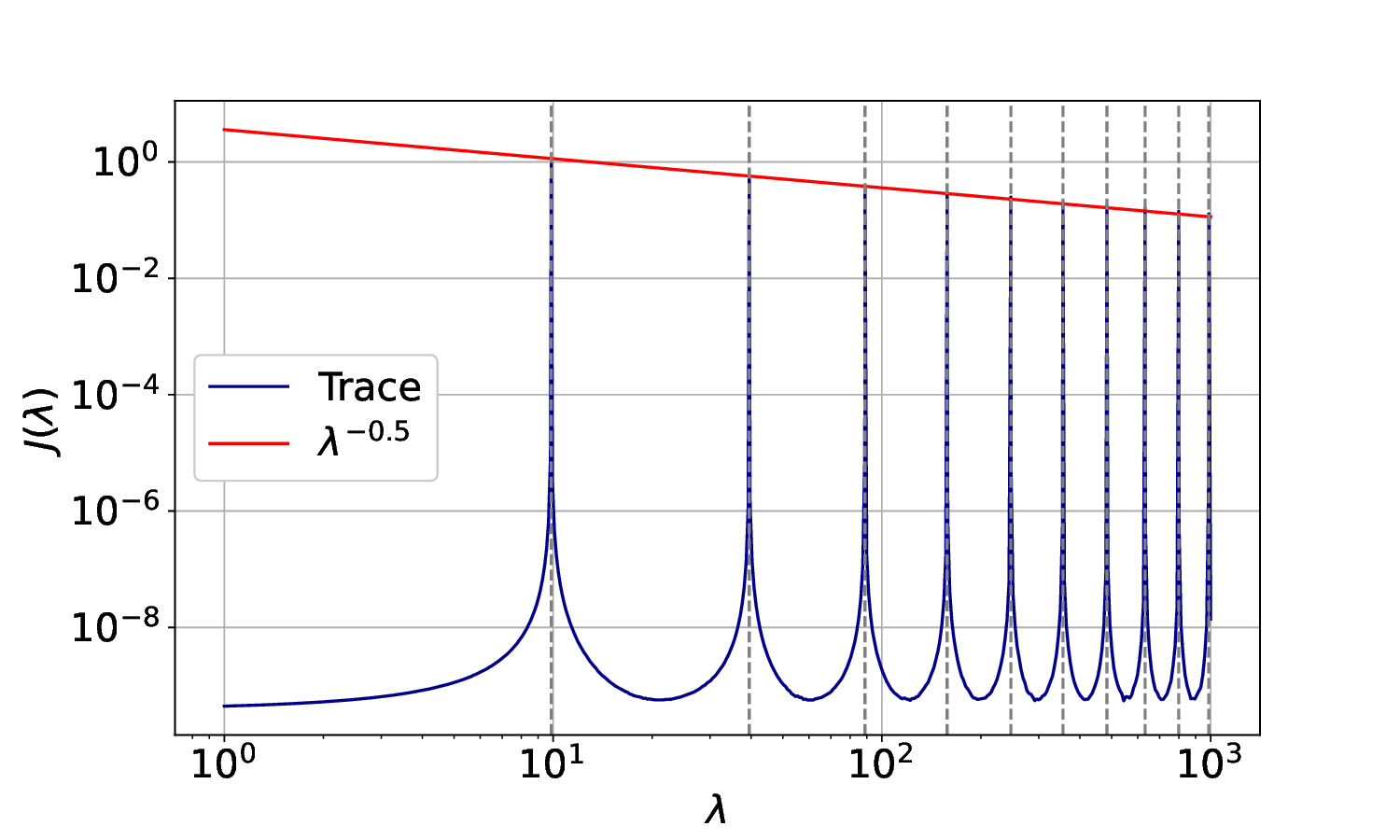}
    \caption{For the 1D Laplace equation, trace of the posterior covariance matrix as a function of $\lambda$. Vertical dashed lines indicate analytical eigenvalue solutions. The red line is the $\lambda^{-0.5}$ trend line.}
    \label{fig:std}
\end{figure}

To apply the PI-GPR framework and Algorithm \ref{alg:eigen_scan_v2} to solve the eigenfunction of the system $(\calL-\lambda I)u = 0 $, we scan the $\lambda$ over $N_{\lambda} = 500$ points, spaced logarithmically on the interval $[1, 1000]$. For the GP prior, we use zero mean and the SE covariance function. The hyperparameters are determined empirically, with the variance fixed at $\sigma^2 = 1$ and the length-scale is prescribed as $l = C N_f^{-1} \lambda^{-0.5}$ (where $C=150$). The choice of $l \propto \lambda^{-0.5}$ is based on the Power Spectral Density (PSD) of the SE covariance function, and a detailed justification for this scaling is provided in the analysis subsequent to the Figure \ref{fig:std}. The factor $C$ is selected without fine-tuning to strike a balance between suppressing high-frequency artefacts and maintaining a well-conditioned covariance matrix for numerical stability. 

To ensure sufficient spatial resolution for validating the theoretical properties of the trace criterion, we employ a relatively large training set of $N=500$ uniform grid collocation points in $[0,1]$. We emphasise that this choice prioritises numerical precision and the clear visualisation of the posterior covariance's behaviour over computational efficiency, as the primary goal here is to provide empirical evidence for Theorem \ref{thm:samp}. The posterior GP is also evaluated on a uniform grid of $N_t = 500$ test points. To maintain numerical stability when the operator $(\mathcal{L}-\lambda)$ approaches singularity, a small jitter term $\eta$ is incorporated into the diagonal of the Gram matrix.

We plot the trace of the posterior covariance across the scanned range of $\lambda$ in Figure \ref{fig:std}, with both axes shown on a logarithmic scale. As $\lambda$ approaches the true eigenvalues (marked by grey dashed lines), the trace exhibits sharp peaks. These local maxima confirm that the posterior covariance effectively characterises the uncertainty inherent in a null space with undetermined amplitude. In the non-eigenvalue regions, the trace drops to a baseline below $10^{-8}$. While our theory suggests a total collapse to zero, the use of finite discretisation ($N=500$) and a small jitter term ($10^{-8}$) for numerical stability precludes an absolute zero in practice. 

In the 1D Laplace eigenvalue problem, each eigenmode corresponds to a specific frequency $\omega_k = \sqrt{\lambda_k}$. While using a GP to approximate the eigenfunctions, its representative power is dictated by the Power Spectral Density (PSD) of the covariance function. The PSD $S(\omega)$ is the Fourier transform of the stationary kernel, representing the distribution of power across the frequency spectrum. For the Squared Exponential (SE) kernel, this manifests as $S(\omega) \propto l \exp(-l^2\omega^2/2)$. Consequently, as the eigenfrequency $\omega_k$ increases, the power assigned to higher-order modes would decay exponentially, as these modes shift into the far-tail region of the Gaussian PSD. To compensate for this attenuation and ensure a consistent representation across scales, we adaptively scale the length-scale as $l \propto \lambda_k^{-0.5}$. This adjustment effectively aligns the bandwidth of the SE covariance function with the characteristic wavelength of each eigenmode, keeping the exponential term in the PSD constant. Under this adaptive scaling, the decay is no longer dominated by the exponential tail but by the pre-factor $l$, yielding the final scaling $J(\lambda_k) \propto l \propto \lambda_k^{-0.5}$. This theoretical alignment explains the decay observed in Figure \ref{fig:std} as a coupling between the PSD of the SE covariance function and the physical frequency of the eigenfunctions.

\begin{figure}[htbp]
    \centering    
    \begin{flushright}
    \includegraphics[height=2.0cm]{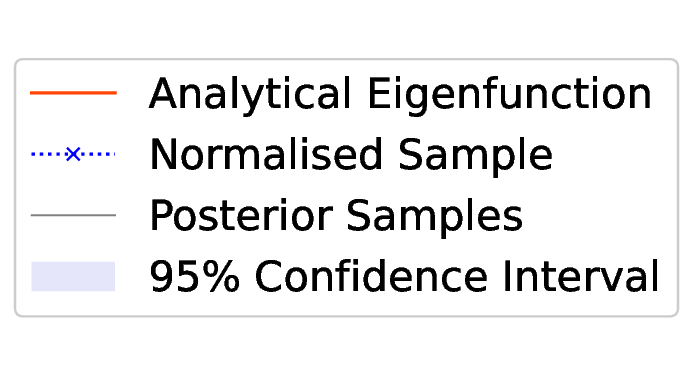} 
    \end{flushright}
    \vspace{-.5cm} 
    \begin{subfigure}[b]{0.4\textwidth}
        \centering        \includegraphics[width=\textwidth]{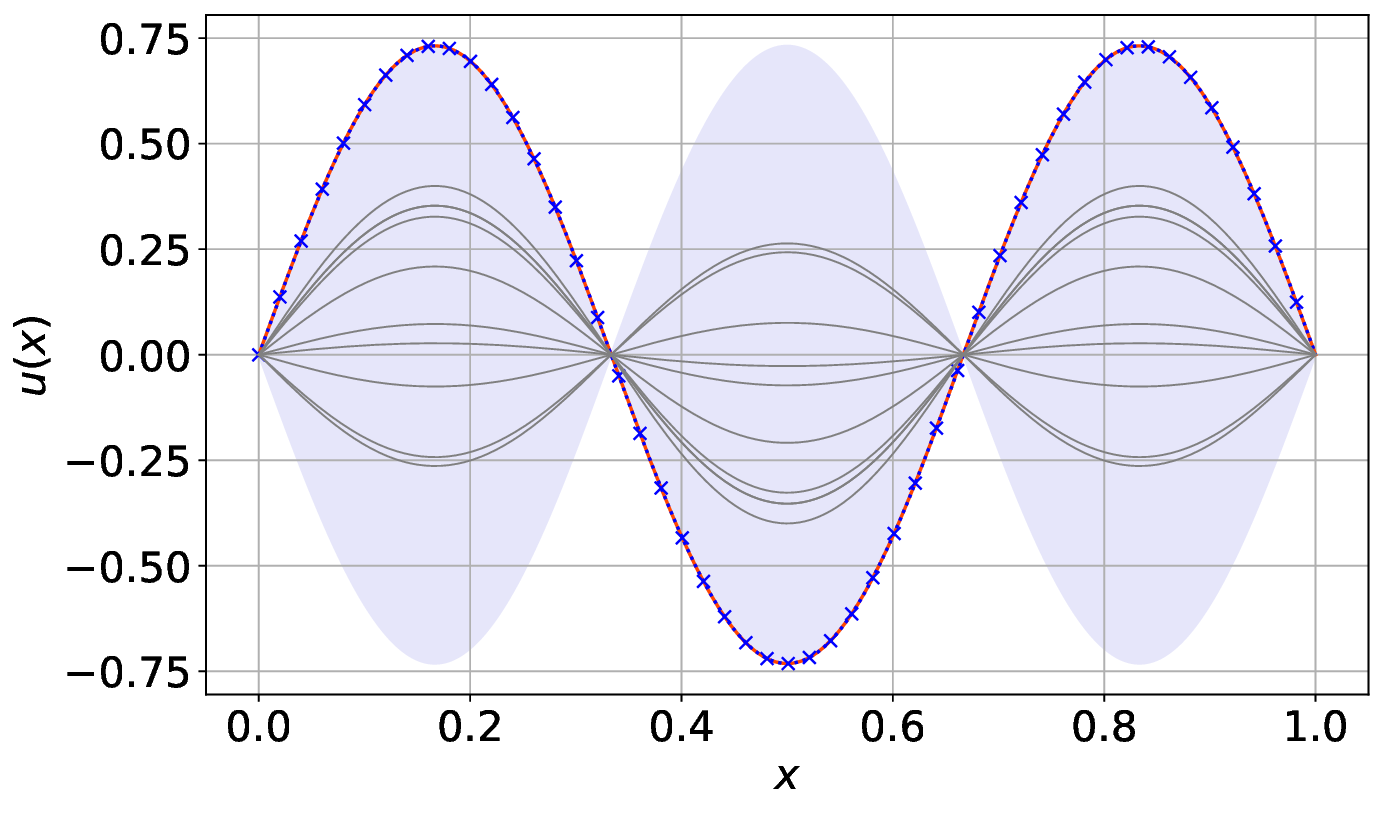}   \renewcommand\thesubfigure{a.1}
        \caption{}
        \label{fig:a1}
    \end{subfigure}
    \begin{subfigure}[b]{0.4\textwidth}
        \centering
        \includegraphics[width=\textwidth]{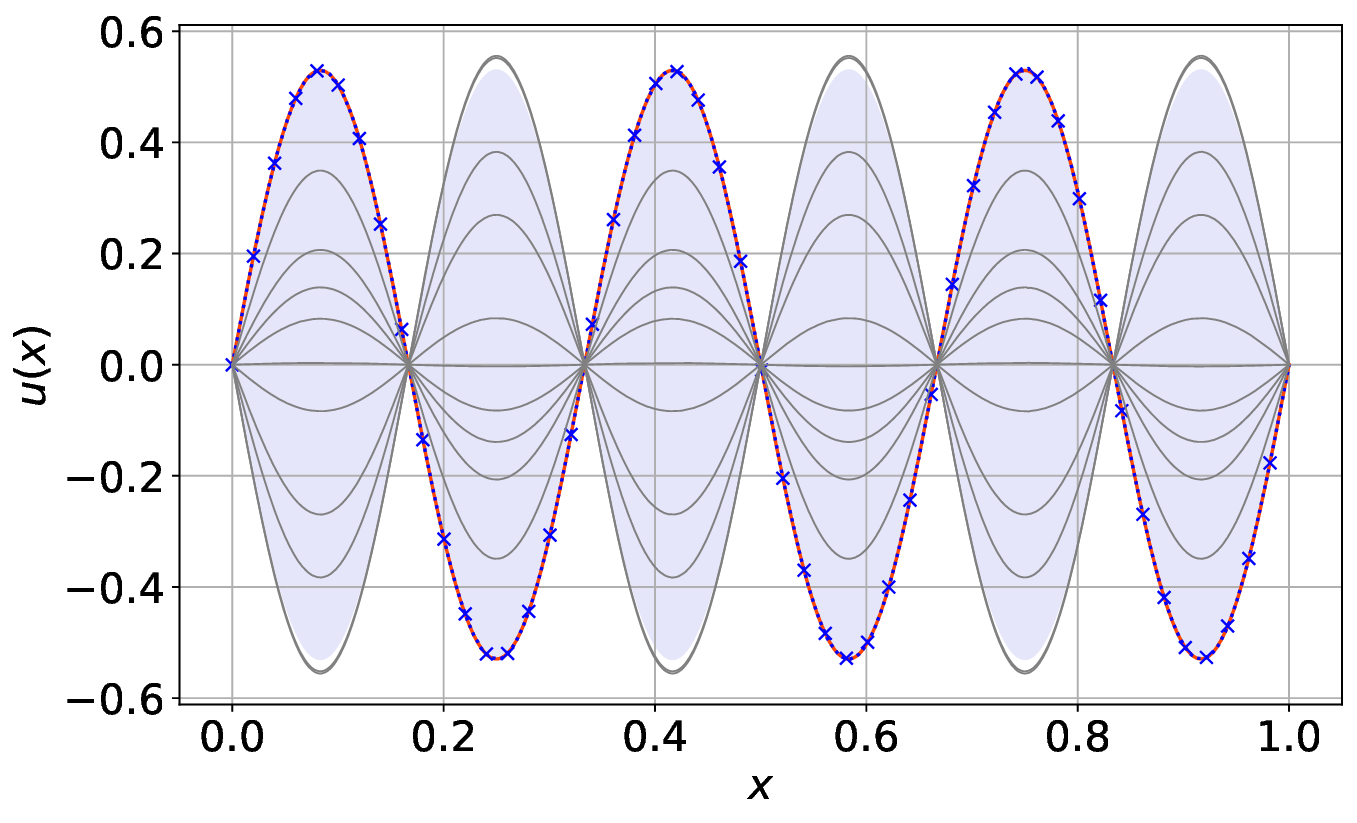}   \renewcommand\thesubfigure{b.1}
        \caption{}
        \label{fig:b1}
    \end{subfigure}    
    \begin{subfigure}[b]{0.4\textwidth}
        \centering        \includegraphics[width=\textwidth]{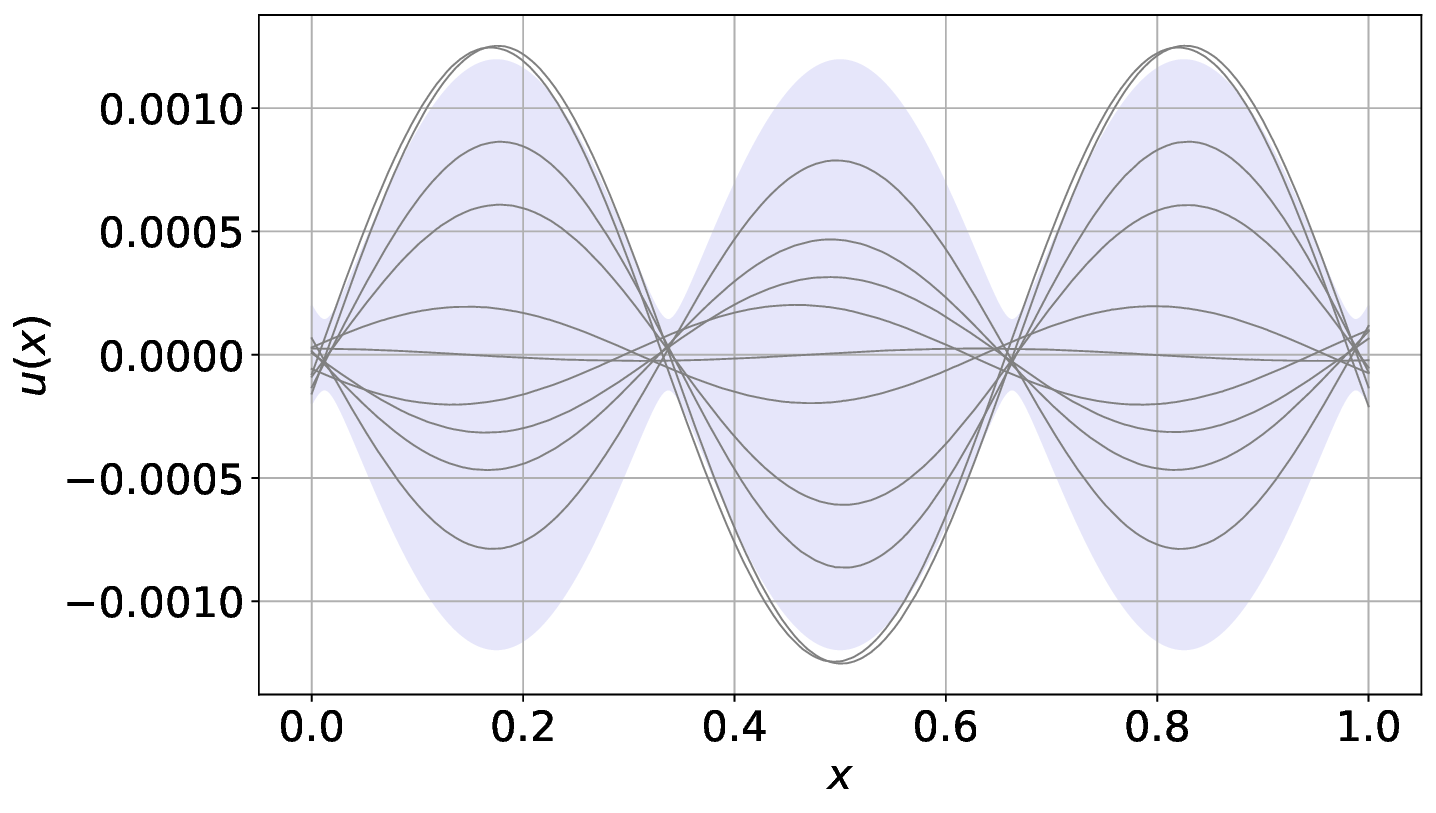}   \renewcommand\thesubfigure{a.2}
        \caption{}
        \label{fig:a2}
    \end{subfigure}
    \begin{subfigure}[b]{0.4\textwidth}
        \centering        \includegraphics[width=\textwidth]{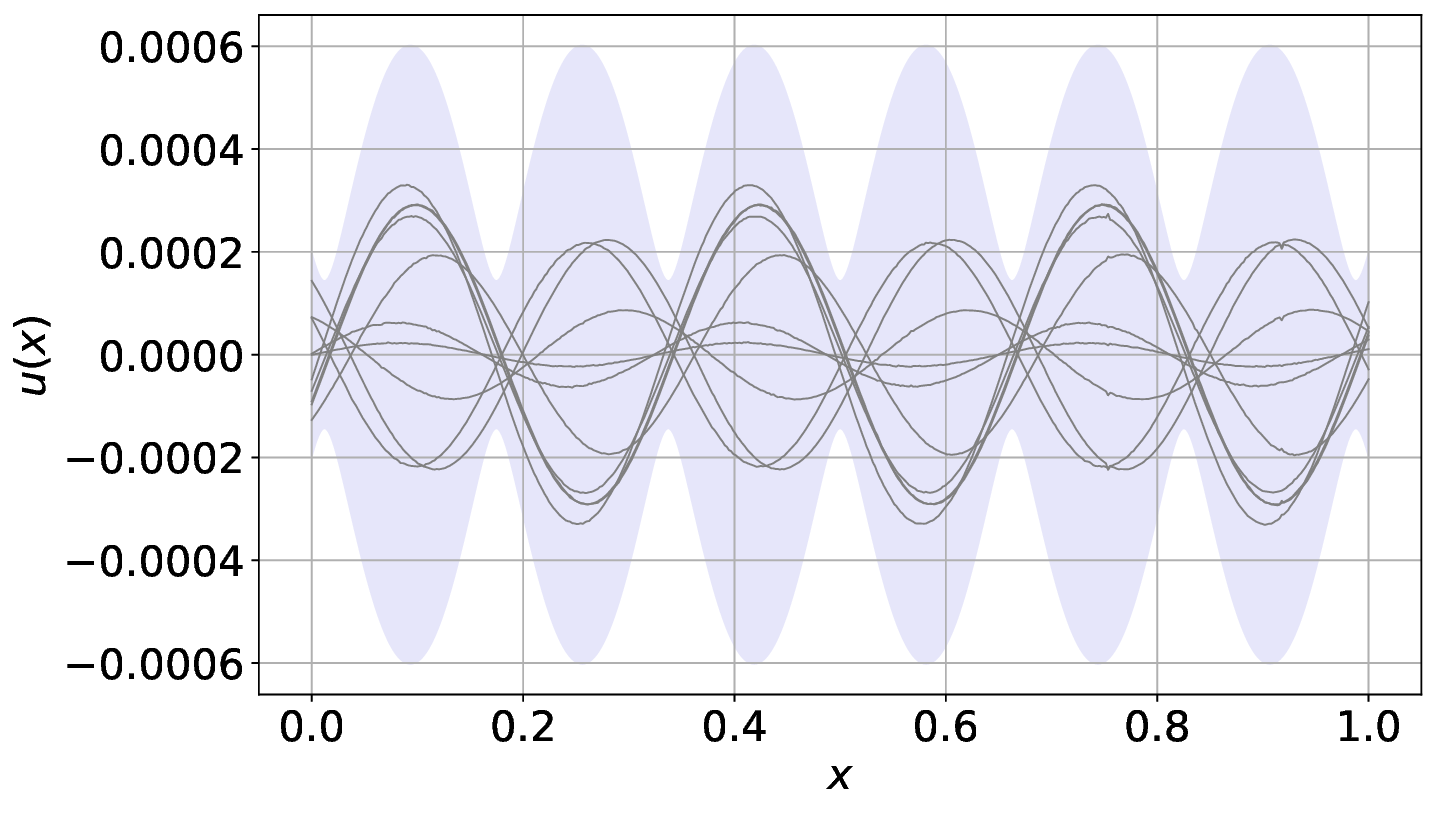}   \renewcommand\thesubfigure{b.2}
        \caption{}
        \label{fig:b2}
    \end{subfigure}
    \begin{subfigure}[b]{0.4\textwidth}
        \centering        \includegraphics[width=\textwidth]{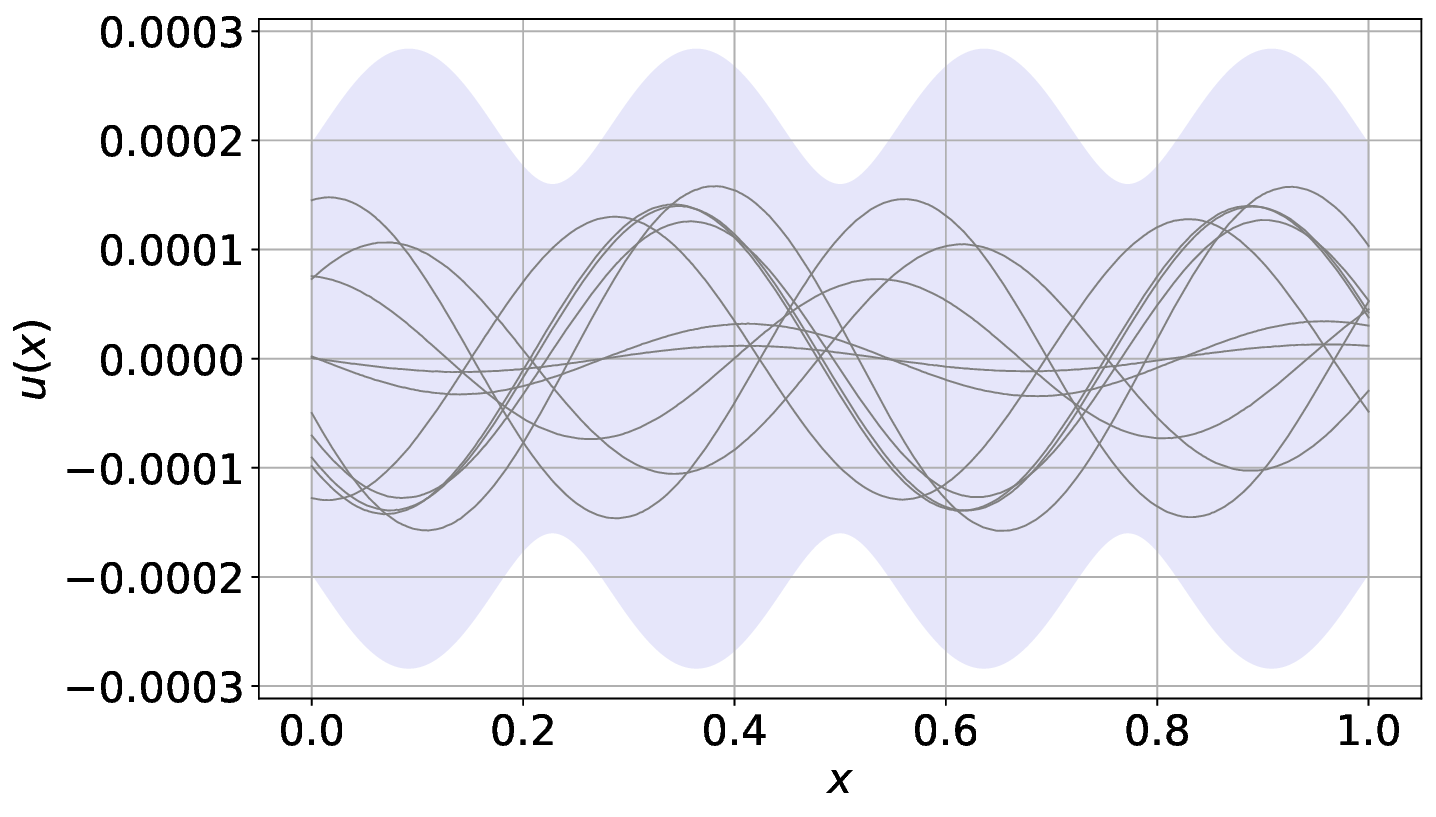}   \renewcommand\thesubfigure{a.3}
        \caption{}
        \label{fig:a3}
    \end{subfigure}
    \begin{subfigure}[b]{0.4\textwidth}
        \centering        \includegraphics[width=\textwidth]{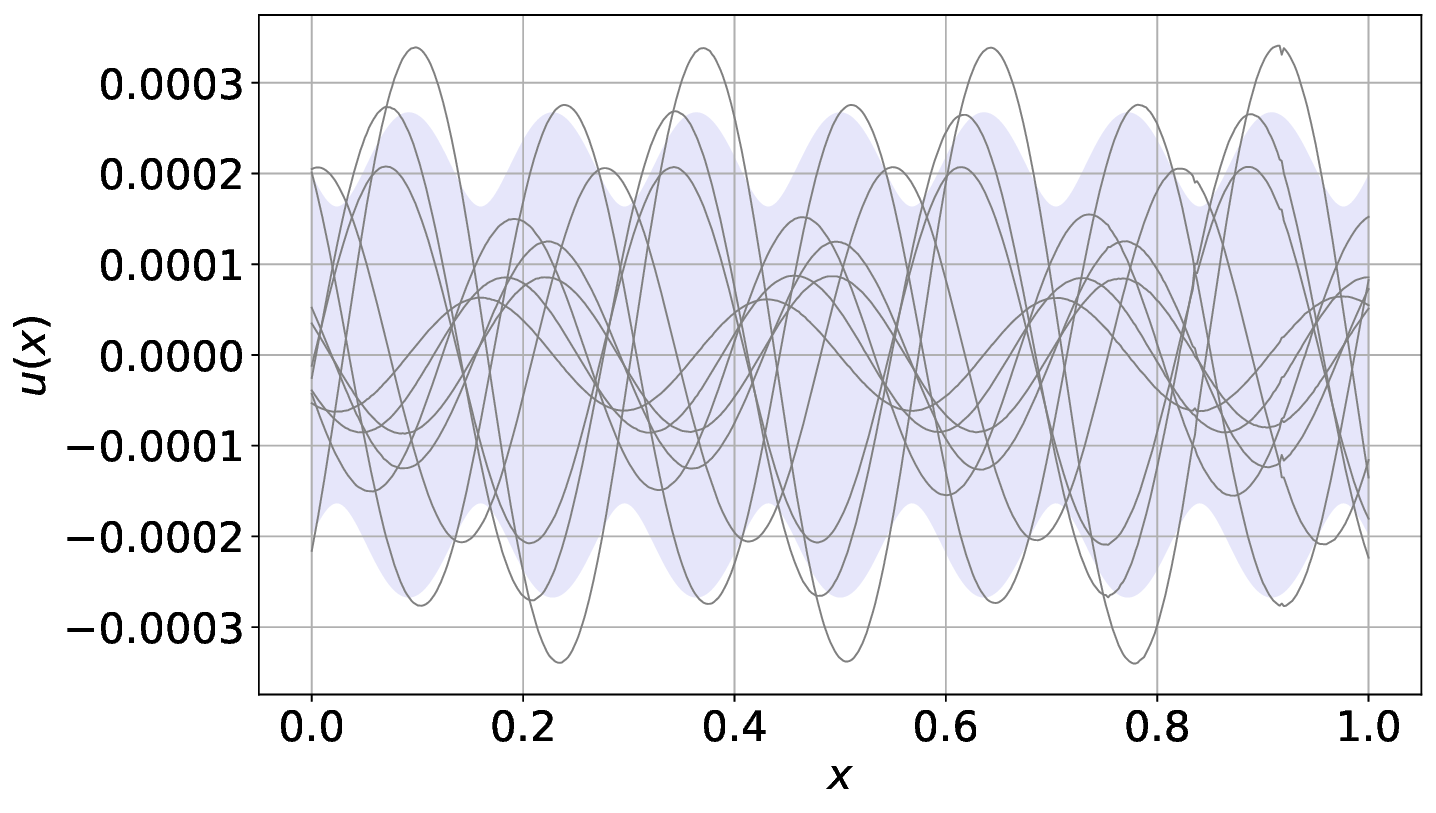}   \renewcommand\thesubfigure{b.3}
        \caption{}
        \label{fig:b3}
    \end{subfigure}

    \caption{Confidence interval, samples and normalised samples of the posteriors of the 1D Laplace problem, for two different $\lambda$s ($\lambda_3$ and $\lambda_6$) and their perturbations: (a.1) $\lambda = \lambda_3$, (a.2) $\lambda = 1.05\lambda_3$, (a.3) $\lambda = 1.5\lambda_3$, (b.1) $\lambda = \lambda_6$, (b.2) $\lambda = 1.05\lambda_6$, (b.3) $\lambda = 1.5\lambda_6$. The corresponding analytical eigenfunctions are drawn in red lines. The shaded regions are the $95\%$ confidence intervals. Grey lines are samples drawn from the posteriors. Note that the scale of the vertical axis varies across different subplots.}
    \label{fig:Helm}
\end{figure}




In Figure \ref{fig:Helm}, we plot the change of the PI-GPR posterior as the candidate $\lambda$ deviates from the true eigenvalues of the PDE. In the first row, we have $\lambda = \lambda_{n}$. In this case, the posterior exhibits a high degree of uncertainty, reflecting the existence of a non-trivial null space with undetermined amplitude. The samples drawn from this distribution exhibit a coherent modal structure that aligns exactly with the analytical eigenfunctions, indicating the PI-GPR posterior solves the eigenvalue equation of the underlying PDE. The second row,  $\lambda_n$ is slightly perturbed, $\lambda = 1.05 \lambda_{n}$). The uncertainty is largely reduced in this case. Notice that although the prior uncertainty is the same across the different subplot cases, the magnitude of the posterior covariance of 2nd row drops to $10^{-4}$, indicating that the operator $(\mathcal{L}-\lambda)$ is becoming invertible. Nevertheless, the samples still retain a recognisable modal shape as the deviation from the true eigenvalues is small ($5\%$). The third Row plots results with $\lambda = 1.5 \lambda_{n}$). It is clear that when $\lambda$ is far away from the true eigenvalues, the posterior covariance is even smaller. While the sample remains periodic, they appear as low-amplitude oscillations with random phases. One reason that they do not vanish completely is due to the $10^{-8}$ diagonal jitter added for numerical stability and finite discretisation effects.

\subsection{One-dimensional cantilevered beam}
In this section, we apply our PI-GPR framework to the eigenvalue problem of a one-dimensional fourth-order operator on $[0,1]$:
\begin{equation}\label{equ:bih}
    \frac{d^{4}u}{dx^{4}} \;-\;\lambda\,u \;=\;0,
    \qquad x\in(0,1),
\end{equation}
subject to the boundary conditions
\begin{equation}\label{equ:bih_bc}
    u(0)=u'(0)=0, 
    \quad\;
    u''(1)=u'''(1)=0.
\end{equation}
Eqs.(\ref{equ:bih})–(\ref{equ:bih_bc}) describe the classical clamped–free cantilever beam eigenvalue problem. The eigenfunctions $\{u_{n}\}$ can be described by a combination of hyperbolic and sinusoidal functions: 
\[
    u(x)=C_{1}\cosh(\alpha x)+C_{2}\sinh(\alpha x)
        +C_{3}\cos(\alpha x)+C_{4}\sin(\alpha x).
\]
where $C_n$ are coefficients depending on the boundary condition and we have denoted $\lambda=\alpha^{4}$. Admissible solutions can be found via the roots of the determinant equation $\cosh\alpha\,\cos\alpha \;=\;1$ and the first few $\alpha_n$ can be found analytically as:
\[
    \alpha_{1} \approx 1.875,\;
    \alpha_{2} \approx 4.730,\;
    \alpha_{3} \approx 7.853,\;
    \alpha_{4} \approx 10.996,\;
    \alpha_{5} \approx 14.137,
    \qquad
    \lambda_{n}=\alpha_{n}^{4}.
\]
Applying the PI-GPR to this example, we evaluate the posterior trace $J(\lambda)$ at $N_{\lambda} = 500$ points, where $\lambda^{1/4}$ is uniformly spaced over the interval $[1, 15]$. Again, we use the zero mean and the SE covariance function GP prior. The hyperparameters are determined empirically, with the variance fixed at $\sigma^2 = 1$ and the length-scale is prescribed as $l = C N_f^{-1} \lambda^{-\frac{1}{4}}$ (where $C=1000$). Analogous to the second-order case described in Section \ref{subsec:2nd order}, in this fourth-order example we use $l \propto \lambda^{-\frac{1}{4}}$ to counteract the exponential decay of the Power Spectral Density. Meanwhile, the fourth-order operator results in eighth-order derivatives of the base covariance function, which significantly increase the numerical sensitivity of the system. Thus, we increase the jitter to $10^{-5}$ to ensure stable matrix inversion. The training and test points are still $N = N_t = 500$ uniform grid collocation points in $[0,1]$.

\begin{figure}[htbp]
    \centering
    \includegraphics[width=0.6\textwidth]{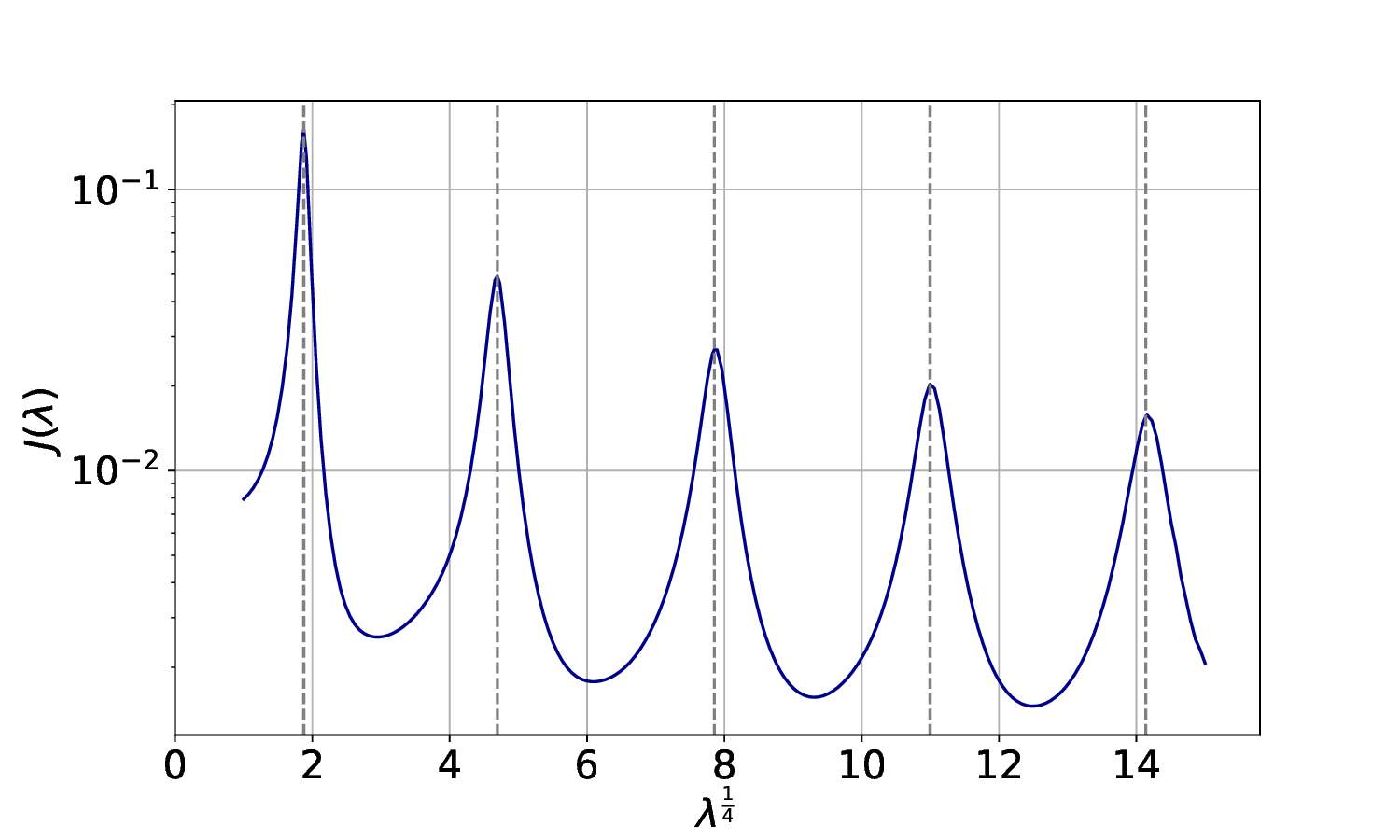}
    \caption{Posterior trace for the cantilever beam example. Same key as Fig. \ref{fig:std}}
    \label{fig:4th}
\end{figure}

\begin{figure}[htbp]
    \centering    
    \begin{flushright}
    \includegraphics[height=2.0cm]{numerics/legend_box.eps} 
    \end{flushright}
    \vspace{-0.5cm} 
    \begin{subfigure}[b]{0.4\textwidth}
        \centering        \includegraphics[width=\textwidth]{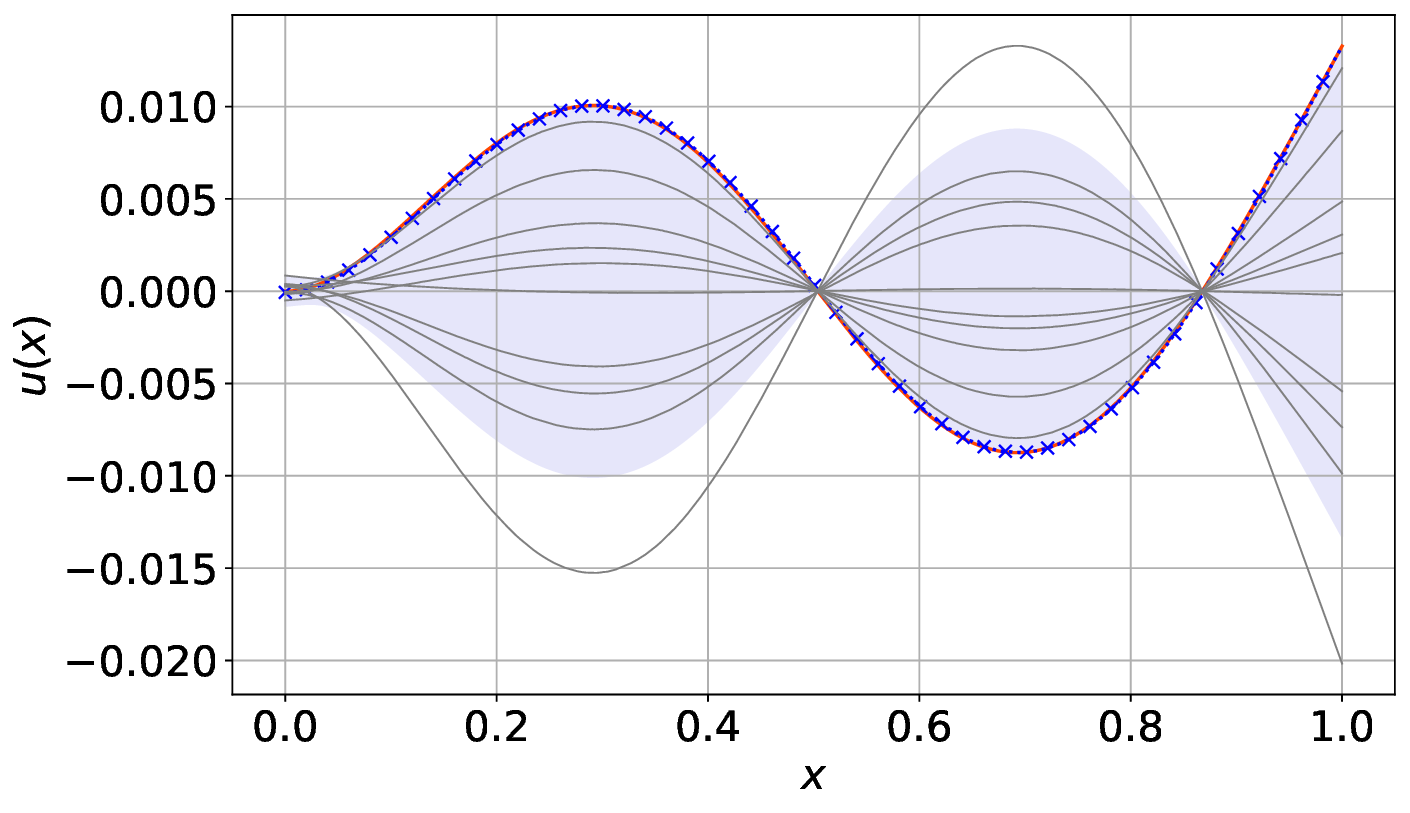}   \renewcommand\thesubfigure{a.1}
        \caption{}
        \label{fig:2a1}
    \end{subfigure}
    \begin{subfigure}[b]{0.4\textwidth}
        \centering
        \includegraphics[width=\textwidth]{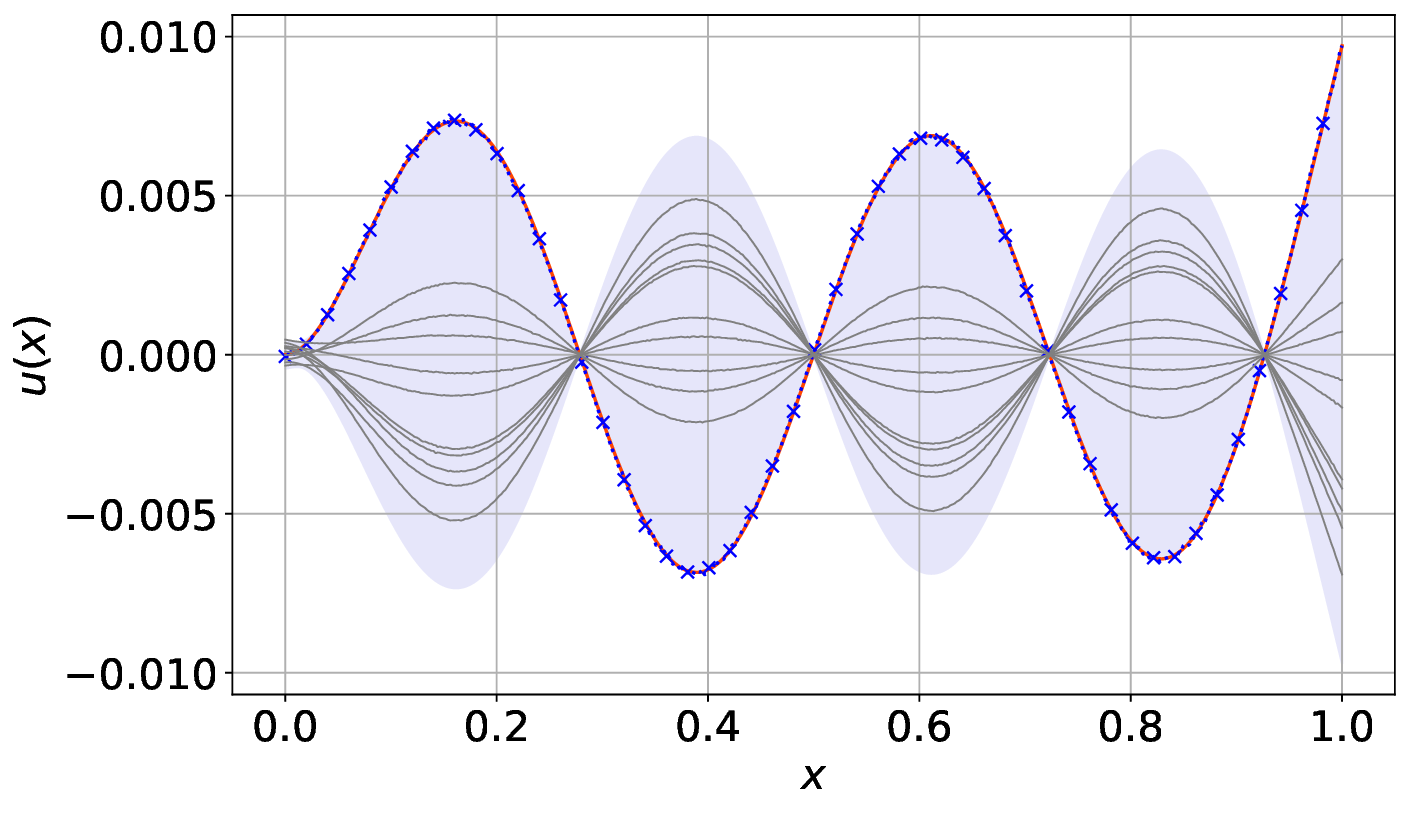}   \renewcommand\thesubfigure{b.1}
        \caption{}
        \label{fig:2b1}
    \end{subfigure}    
    \begin{subfigure}[b]{0.4\textwidth}
        \centering        \includegraphics[width=\textwidth]{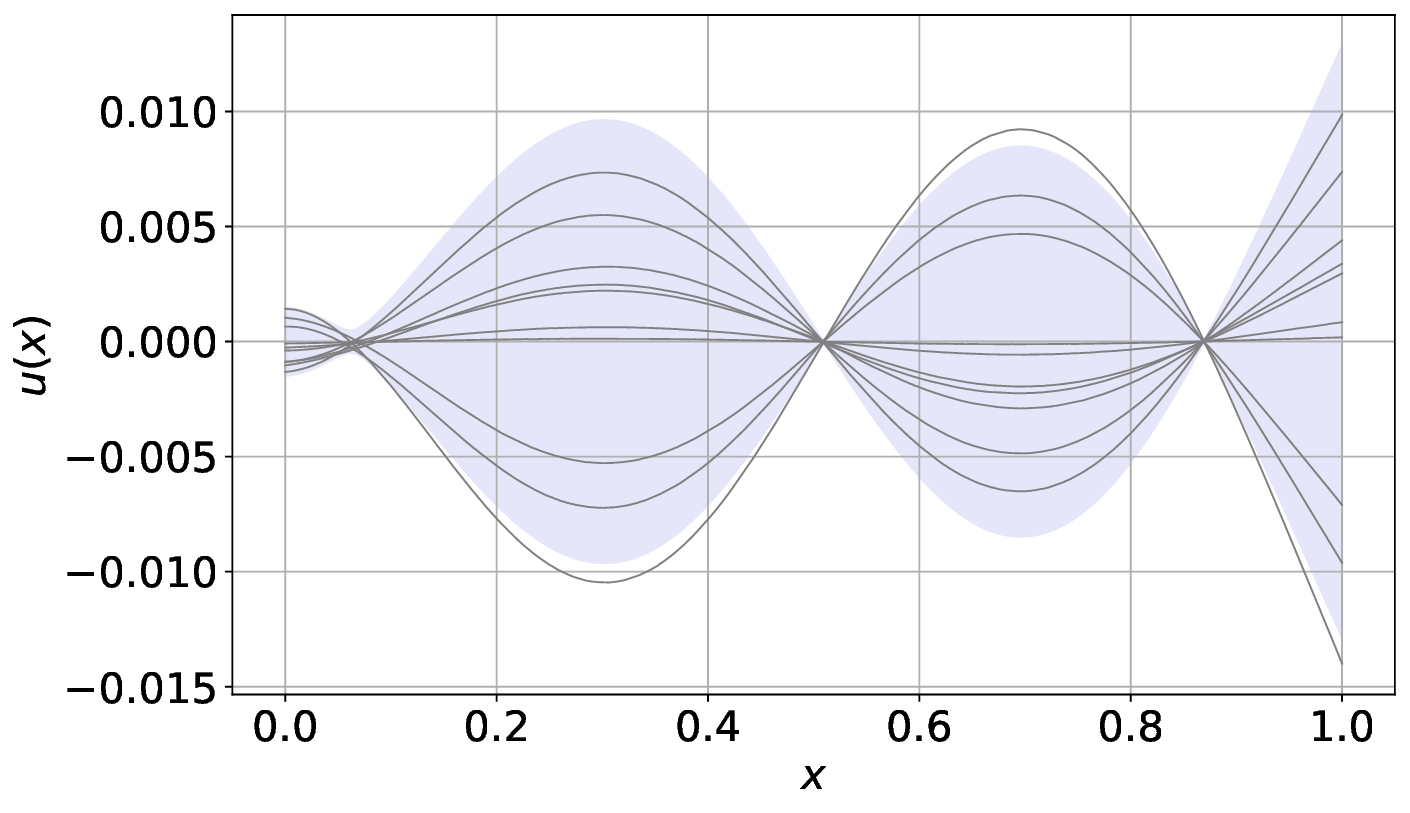}   \renewcommand\thesubfigure{a.2}
        \caption{}
        \label{fig:2a2}
    \end{subfigure}
    \begin{subfigure}[b]{0.4\textwidth}
        \centering        \includegraphics[width=\textwidth]{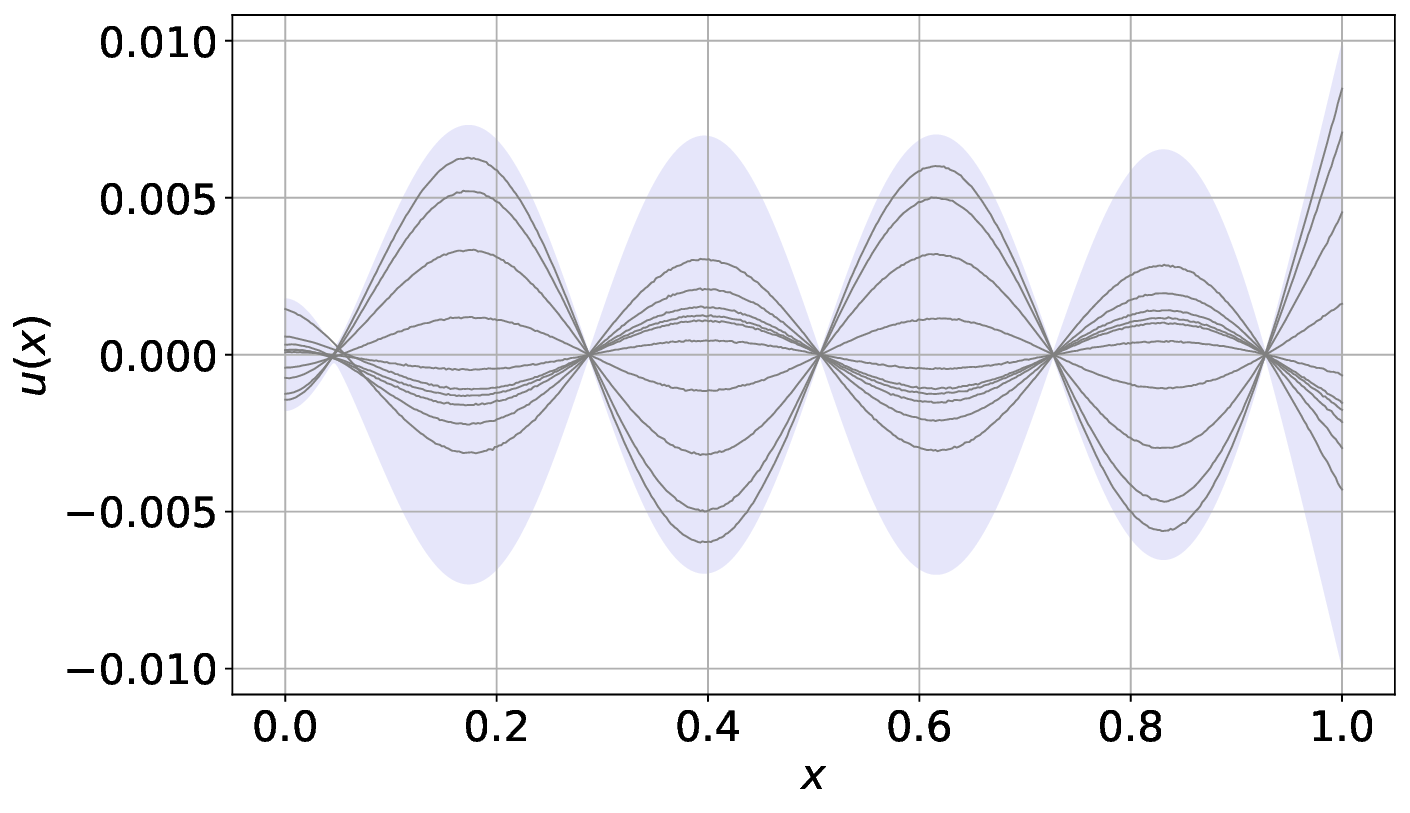}   \renewcommand\thesubfigure{b.2}
        \caption{}
        \label{fig:2b2}
    \end{subfigure}
    \begin{subfigure}[b]{0.4\textwidth}
        \centering        \includegraphics[width=\textwidth]{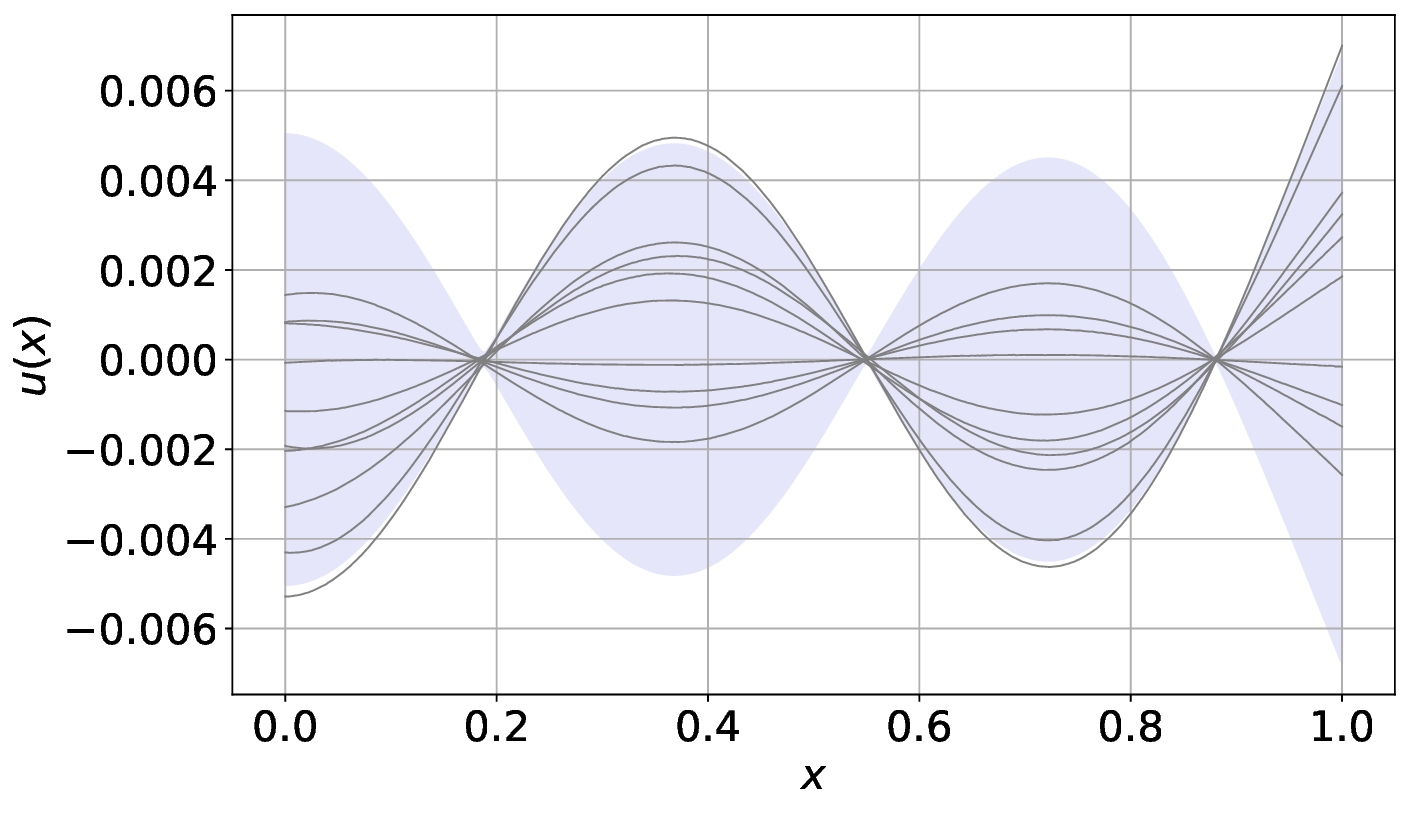}   \renewcommand\thesubfigure{a.3}
        \caption{}
        \label{fig:2a3}
    \end{subfigure}
    \begin{subfigure}[b]{0.4\textwidth}
        \centering        \includegraphics[width=\textwidth]{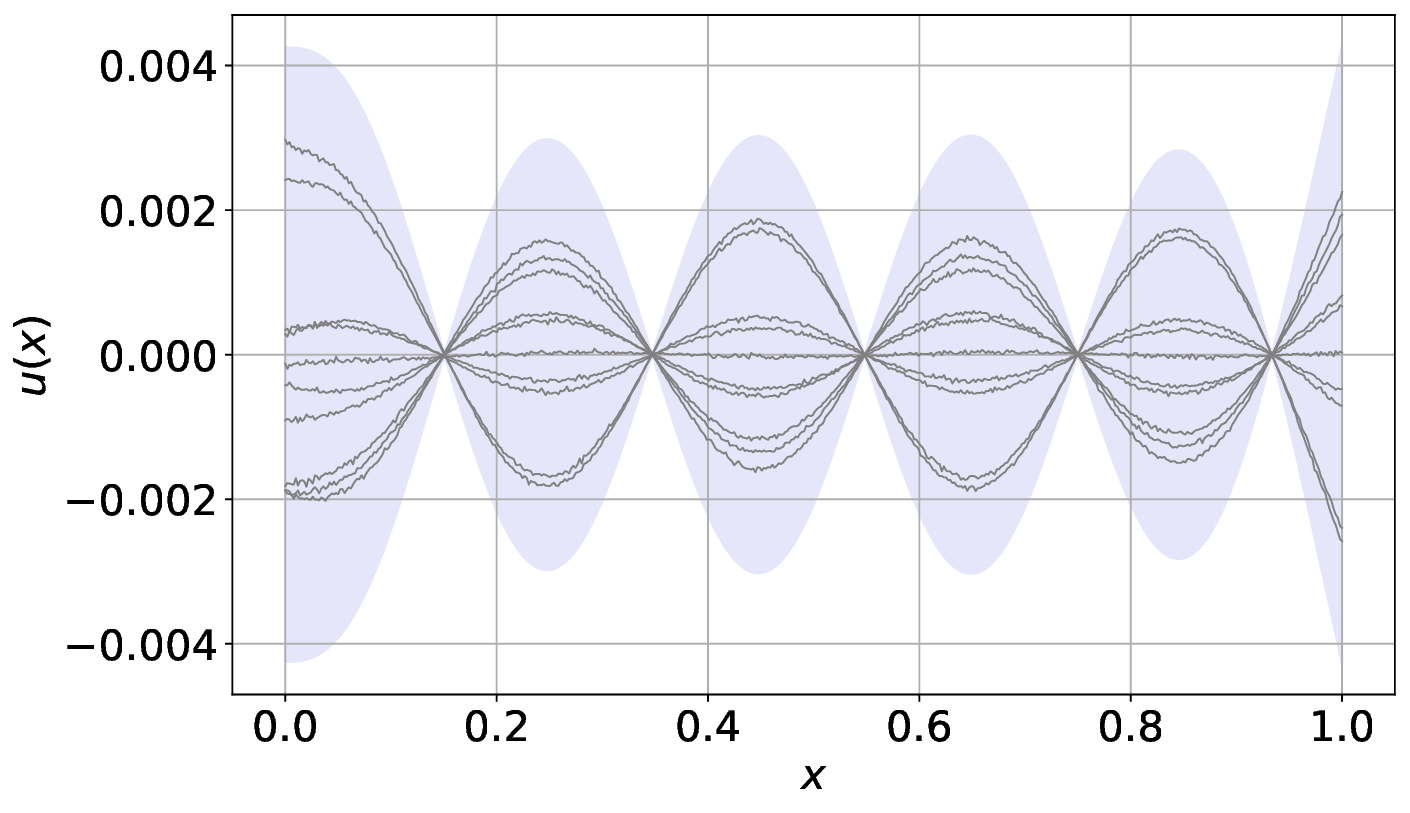}   \renewcommand\thesubfigure{b.3}
        \caption{}
        \label{fig:2b3}
    \end{subfigure}

    \caption{Confidence interval, samples and normalised samples of the posteriors of the cantilever beam example, for different $\lambda$s: (a.1) $\lambda = \lambda_2$, (a.2) $\lambda = 1.05\lambda_2$, (a.3) $\lambda = 1.5\lambda_2$, (b.1) $\lambda = \lambda_4$, (b.2) $\lambda = 1.05\lambda_4$, (b.3) $\lambda = 1.5\lambda_4$. The corresponding analytical eigenfunctions are drawn in red lines. The shaded regions are the $95\%$ confidence intervals. Grey lines are samples drawn from the posteriors.}
    \label{fig:Cant}
\end{figure}

We plot the trace of the posterior covariance along the $\lambda$s in Figure \ref{fig:4th} and posterior samples with perturbed $\lambda$s in Figure \ref{fig:Cant}. We see that when $\lambda$ is close to the real eigenvalues (the grey dashed lines), the trace of the posterior covariance reaches the peak. While the overall trend is similar to the previous experiment, the fourth-order nature of the beam operator and the large jitter ($10^{-5}$) introduce a more gradual uncertainty collapse, characterised by phase shift rather than immediate magnitude suppression (see Figure \ref{fig:Cant}).

\subsection{Loaded String with non-linear boundary condition}

We then consider a non-linear eigenvalue problem from the NLEVP collection \cite{betcke2013nlevp}, which models the eigenvibrations of a loaded string. The NEP is given by
\begin{equation} \label{eq:4.6}
    -\frac{d^2 u}{dx^2} = \lambda u, \quad u(0) = 0, \quad \frac{du}{dx}(1) + \frac{\lambda \kappa M}{\lambda - \kappa}u(1) = 0.
\end{equation}
It models the vibrations of a string with a load of mass $M$ attached to an elastic spring with stiffness $\kappa$. We use the default parameters $M = \kappa = 1$. 
\begin{figure}[htbp]
    \centering
    \includegraphics[width=0.6\textwidth]{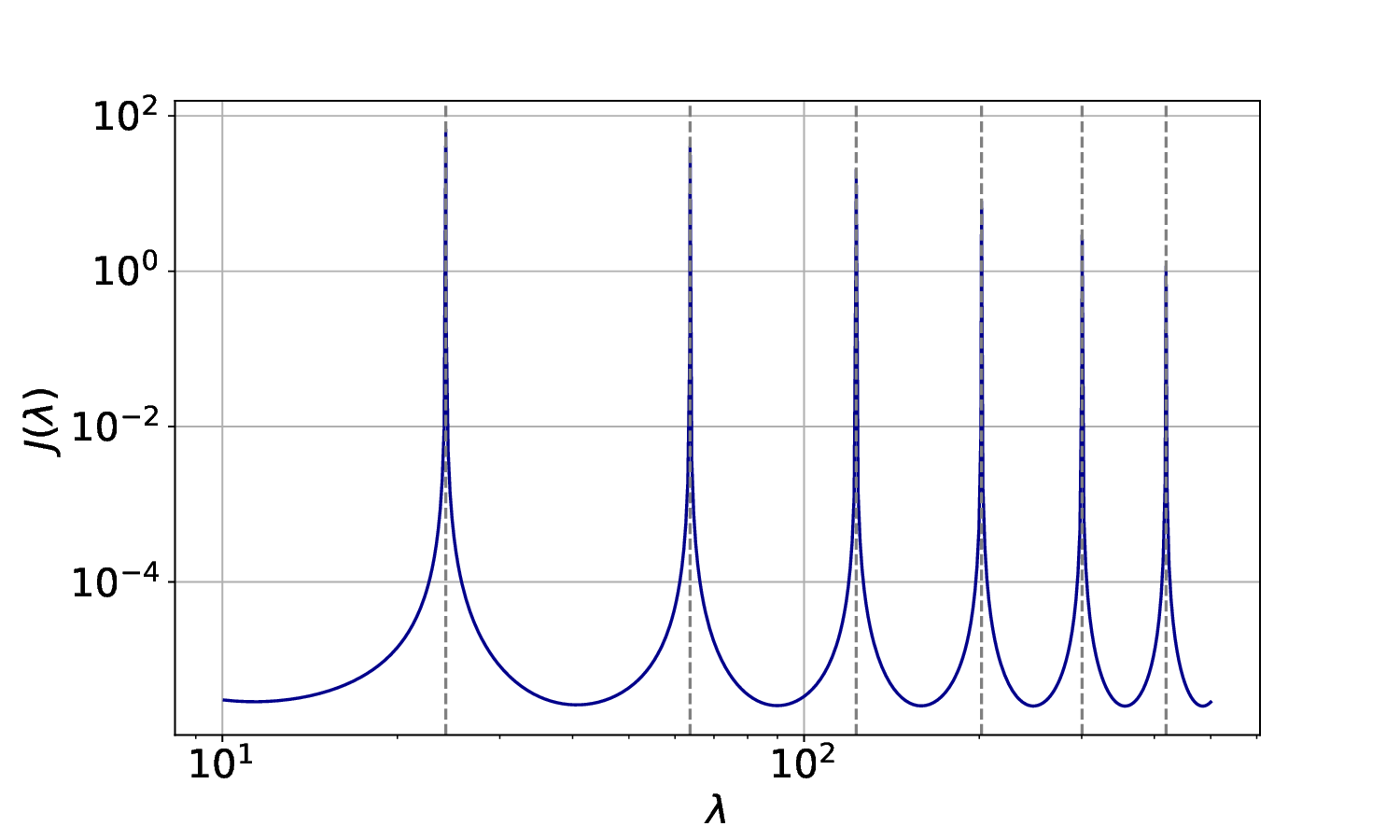}
    \caption{Posterior trace for the loaded string non-linear eigenvalue example. Same key as Fig. \ref{fig:std}}
    \label{fig:lst}
\end{figure}

\begin{figure}[htbp]
    \centering    
    \begin{flushright}
    \includegraphics[height=2.0cm]{numerics/legend_box.eps} 
    \end{flushright}
    \vspace{-0.5cm} 
    \begin{subfigure}[b]{0.4\textwidth}
        \centering        \includegraphics[width=\textwidth]{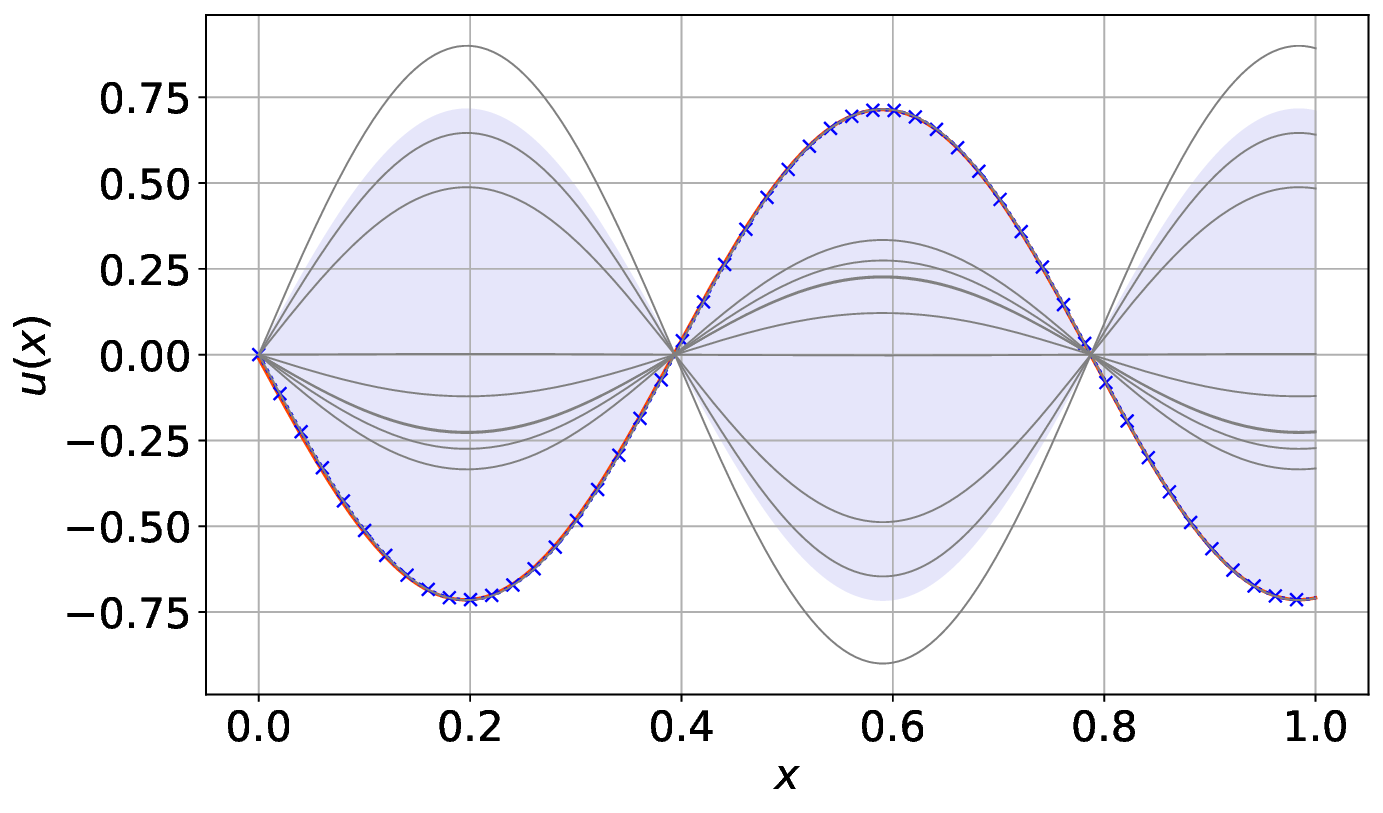}   \renewcommand\thesubfigure{a.1}
        \caption{}
        \label{fig:3a1}
    \end{subfigure}
    \begin{subfigure}[b]{0.4\textwidth}
        \centering
        \includegraphics[width=\textwidth]{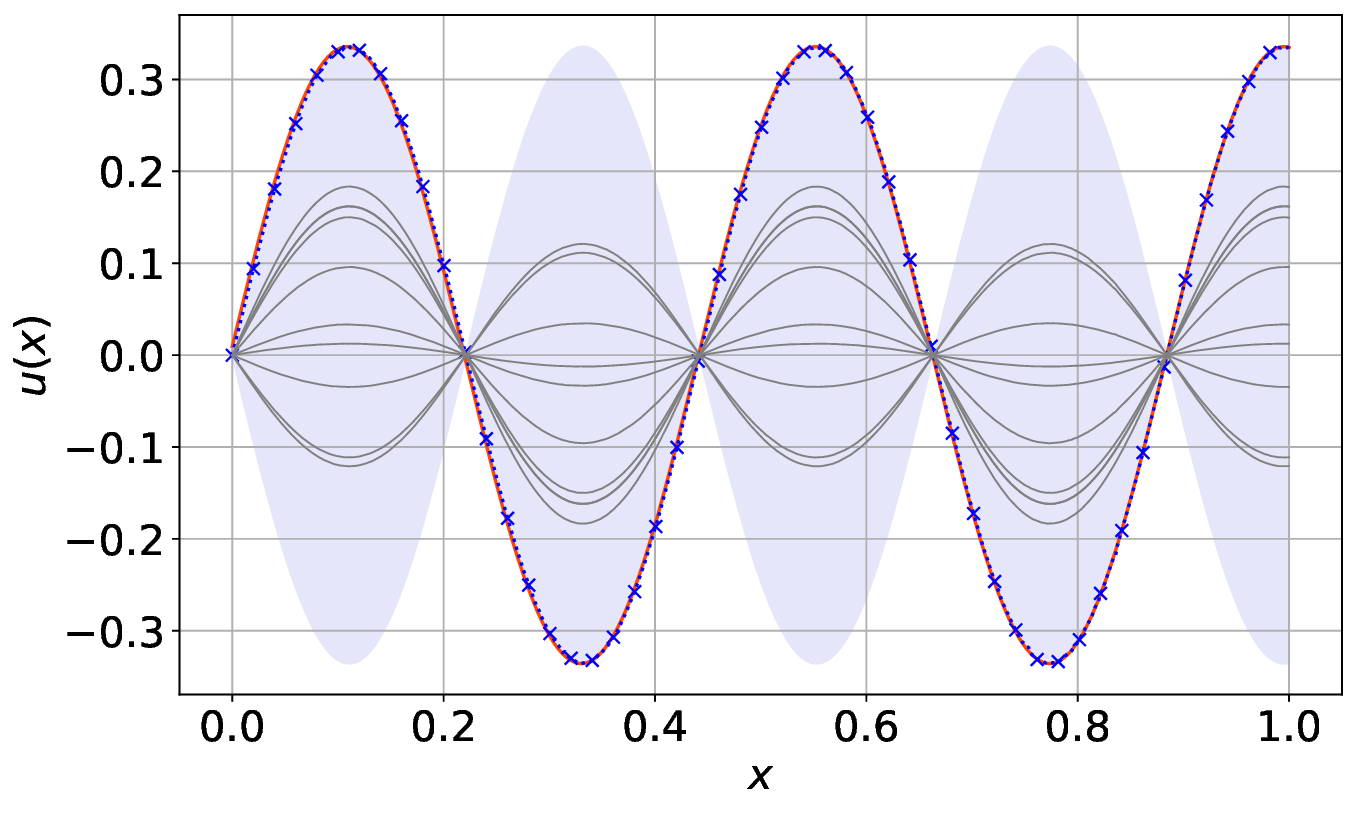}   \renewcommand\thesubfigure{b.1}
        \caption{}
        \label{fig:3b1}
    \end{subfigure}    
    \begin{subfigure}[b]{0.4\textwidth}
        \centering        \includegraphics[width=\textwidth]{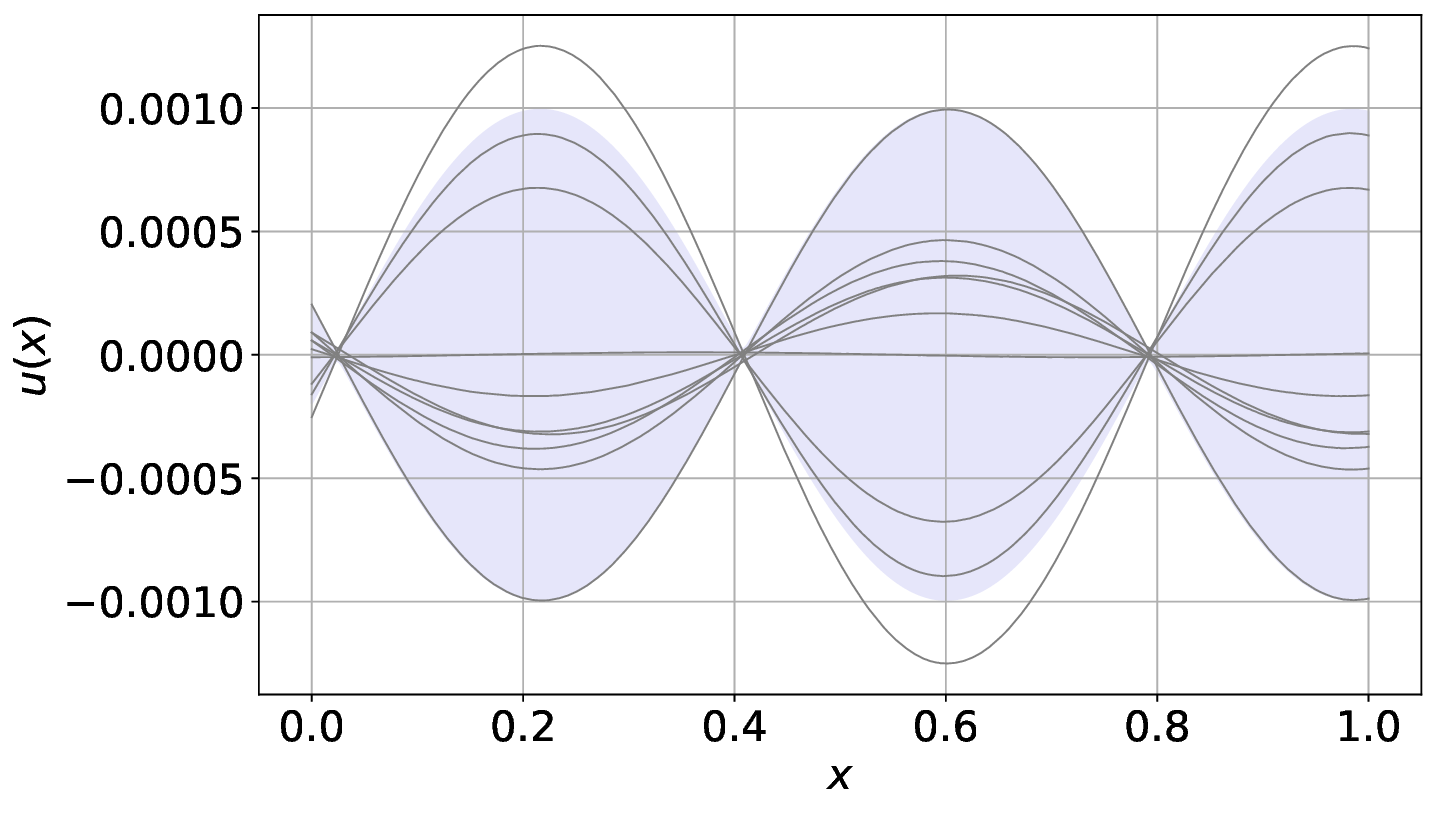}   \renewcommand\thesubfigure{a.2}
        \caption{}
        \label{fig:3a2}
    \end{subfigure}
    \begin{subfigure}[b]{0.4\textwidth}
        \centering        \includegraphics[width=\textwidth]{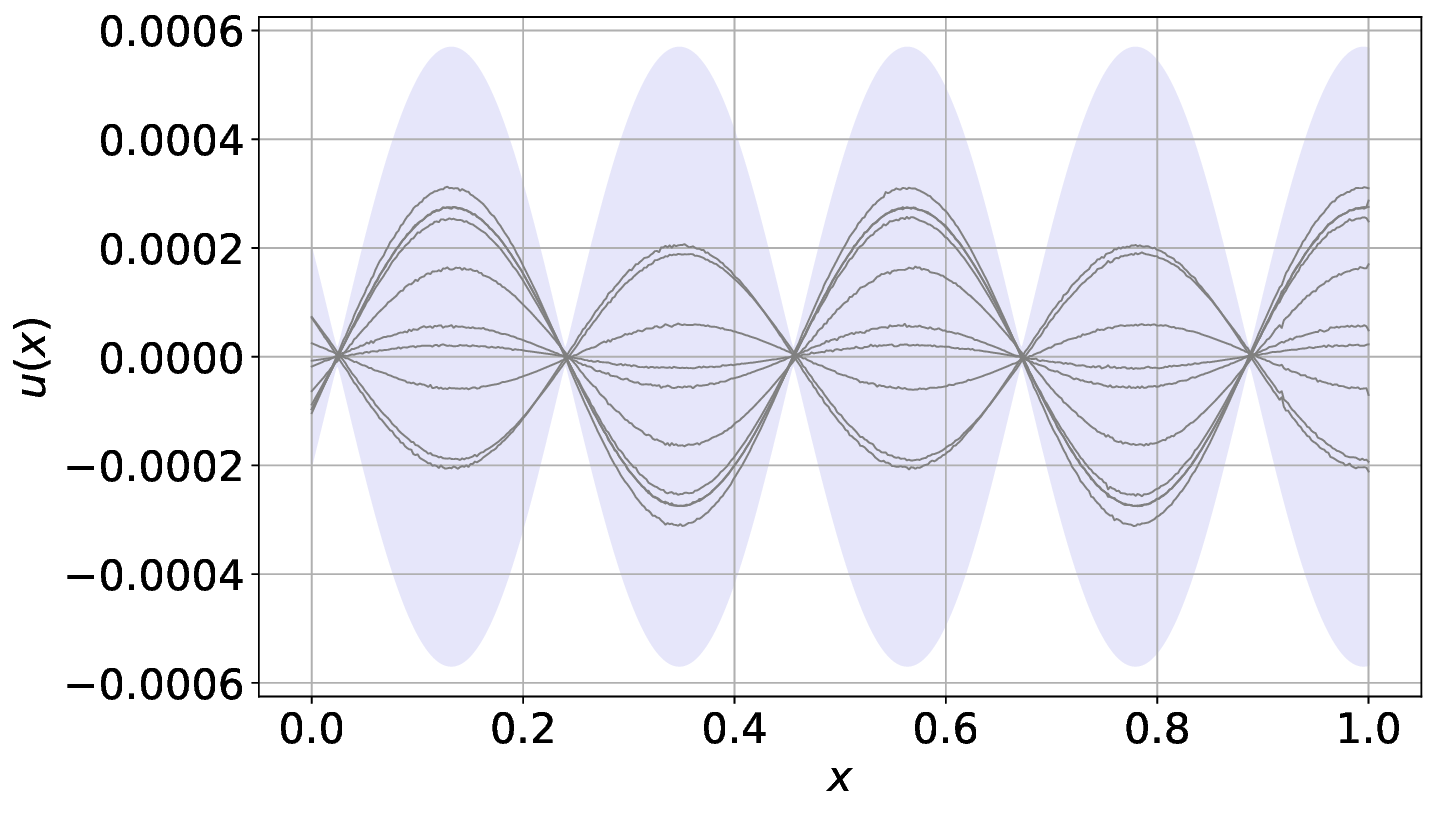}   \renewcommand\thesubfigure{b.2}
        \caption{}
        \label{fig:3b2}
    \end{subfigure}
    \begin{subfigure}[b]{0.4\textwidth}
        \centering        \includegraphics[width=\textwidth]{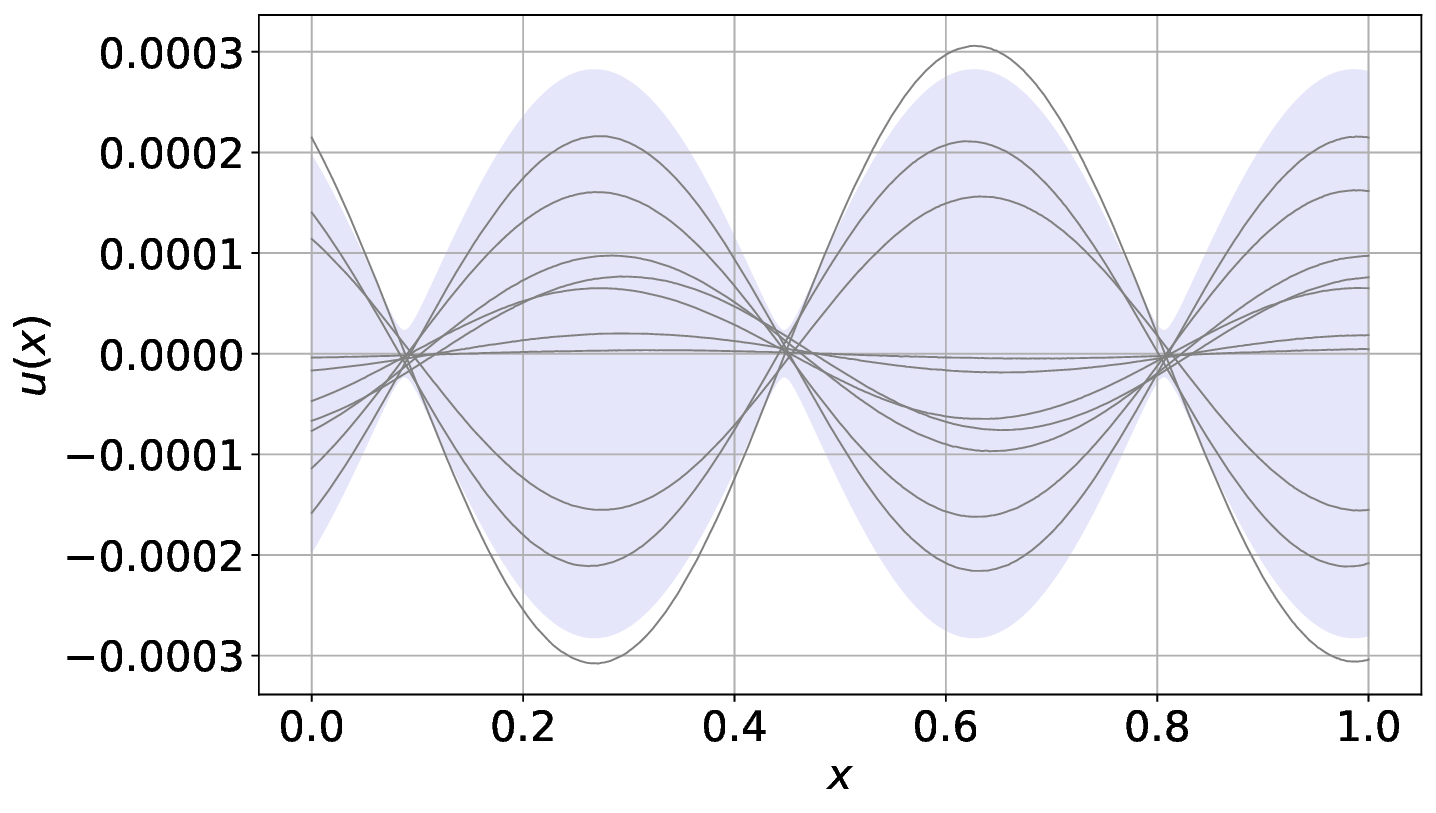}   \renewcommand\thesubfigure{a.3}
        \caption{}
        \label{fig:3a3}
    \end{subfigure}
    \begin{subfigure}[b]{0.4\textwidth}
        \centering        \includegraphics[width=\textwidth]{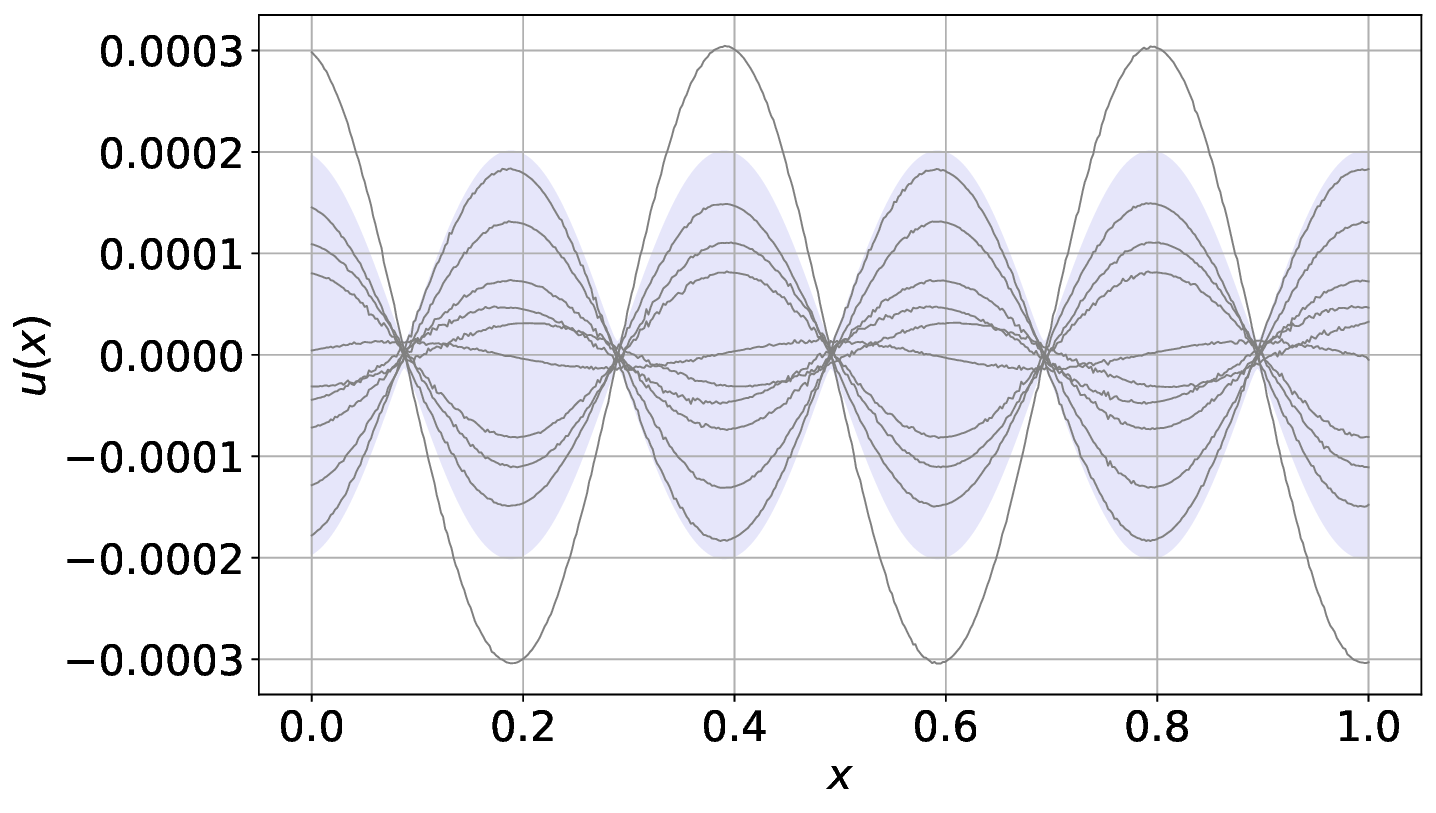}   \renewcommand\thesubfigure{b.3}
        \caption{}
        \label{fig:3b3}
    \end{subfigure}

    \caption{Confidence interval, samples and normalised samples of the posteriors for the loaded string example, for different $\lambda$s: (a.1) $\lambda = \lambda_2$, (a.2) $\lambda = 1.05\lambda_2$, (a.3) $\lambda = 1.5\lambda_2$, (b.1) $\lambda = \lambda_4$, (b.2) $\lambda = 1.05\lambda_4$, (b.3) $\lambda = 1.5\lambda_4$. The corresponding analytical eigenfunctions are drawn in red lines. The shaded regions are the $95\%$ confidence intervals. Grey lines are samples drawn from the posteriors.}
    \label{fig:LoadS}
\end{figure}

We scanned the $\lambda$ over $N_{\lambda} = 500$ points, spaced logarithmically on the interval $[10, 500]$. We adopt the same GP prior as in the first example, which is a zero-mean Gaussian process with an SE covariance function. The hyperparameters are set empirically to $\sigma^2 = 1$ and $l = C N_f^{-1} \lambda^{-1/2}$, where $C = 150$. The jitter is $10^{-8}$ and The training and test points are $N = N_t = 500$ uniform grid collocation points in $[0,1]$. 

We plot the trace of the posterior covariance along the $\lambda$s in Figure \ref{fig:lst}. Both x-axis and y-axis are set in log-scale. We see that when $\lambda$ is close to the real eigenvalues (the grey dashed lines), the trace of the posterior covariance reaches the peak. The $J(\lambda)$ for the loaded string (Figure \ref{fig:LoadS}) is similar to the results observed in the Laplacian experiment. As $\lambda$ deviates from the eigenvalues, we observe a rapid covariance collapse and phase shift.

The inclusion of the loaded string problem serves to highlight an advantage of our proposed framework, which is the seamless handling of non-linear eigenvalue problems (NEPs). In standard numerical methods, discretising such problems yields matrix equations that are non-linear in the eigenvalue parameter $\lambda$. Solving these typically requires specialised treatments, such as reformulation via companion matrices or the use of iterative Newton-type solvers \cite{betcke2013nlevp}. In contrast, our covariance-based approach remains structurally invariant. Since the method treats $\lambda$ as a scanned parameter, the non-linearity in the boundary condition is naturally absorbed into the operator definition at each evaluation step. The GP only requires the operator to be linear with respect to the function $u$, not the eigenvalue $\lambda$. Consequently, our method offers a unified statistical criterion that identifies eigenvalues with equal efficacy, regardless of whether they enter the governing equations linearly or non-linearly.

\section{Conclusion}
\label{sec:conclusion}

In this paper, we have presented a new statistical framework for solving PDE eigenvalue problems using PI-GPR. We addressed the fundamental challenge posed by the homogeneous form of the problem, $(\mathcal{L}-\lambda)u = 0$, where the lack of non-zero data renders the standard posterior mean and the NLML uninformative.

Our core contribution is to shift the focus from the posterior mean to the posterior covariance. We demonstrated that the "size" of the posterior covariance, quantified by its trace $J(\lambda)$, serves as a robust criterion for identifying eigenvalues. When $\lambda$ is not an eigenvalue, the operator admits only the trivial solution, and the posterior covariance collapses to a near-zero matrix. When $\lambda$ is an eigenvalue, the operator admits a non-trivial eigenspace, which is reflected as a sharp peak in the covariance trace.

Furthermore, we showed that at these peaks, the non-zero posterior covariance matrix $\mathbf{K}_{N}(\lambda)$ defines a valid Gaussian measure over the eigenspace. This allows for the approximation of non-trivial eigenfunctions by sampling from the posterior distribution $\mathcal{N}(0, \mathbf{K}_{N}(\lambda))$. We also provided a theoretical explanation for the observed power-law decay of the trace peaks by linking the spectral properties of the covariance function to the physical wavelength of the eigenfunctions.

As a meshless, probabilistic method, our approach offers significant flexibility. For instance, the accuracy and efficiency of the eigenfunction approximation can be enhanced by incorporating constraints at known nodal points ($u(x)=0$), a refinement that is non-trivial to implement in traditional mesh-based methods.

The primary limitation of the current algorithm is its computational cost, which scales with the number of scan points $N_\lambda$ and the $N^3$ cost of matrix inversion. Future work will focus on mitigating this cost, for example by integrating sparse GP approximations to reduce the matrix complexity. In addition, Green’s function-based kernels could potentially provide better resolution for high-order eigenvalues with fewer training points. The scanning process itself could be accelerated by building a GP emulator for the trace function $J(\lambda)$, allowing for Bayesian optimisation to find the peaks efficiently. Further research is also needed to develop a principled method for covariance function hyperparameter optimisation in this zero-data regime, as well as extending this covariance-based framework to non-linear and generalised eigenvalue problems. 


\newpage
\bibliographystyle{unsrt}  
\bibliography{references}

\appendix

\newpage
\section{Proof of Theorem \ref{thm:samp}}
\label{appendix:theorem}

\paragraph{$\lambda$ is Not an eigenvalue of $L$:} If $\lambda$ is not an eigenvalue of $L$, then the matrix $A(\lambda) = L - \lambda I$ is invertible. Since $K$ is symmetric positive definite, it is also invertible. Consequently, the product $A(\lambda)KA(\lambda)^\top$ is invertible. For any invertible matrix, its Moore-Penrose pseudoinverse is equal to its inverse. Therefore, we can write
$$(A(\lambda)KA(\lambda)^\top)^{+} = (A(\lambda)KA(\lambda)^\top)^{-1}$$Using the property that $(XY)^{-1} = Y^{-1}X^{-1}$ for invertible matrices $X, Y$, we have$$(A(\lambda)KA(\lambda)^\top)^{-1} = (A(\lambda)^\top)^{-1}K^{-1}A(\lambda)^{-1}.$$
Substituting this into the expression for $K_N(\lambda)$:
\begin{align*}
K_{N}(\lambda) &= K - KA(\lambda)^\top \left[ (A(\lambda)^\top)^{-1}K^{-1}A(\lambda)^{-1} \right] A(\lambda)K \\
&= K - K \left[ A(\lambda)^\top (A(\lambda)^\top)^{-1} \right] K^{-1} \left[ A(\lambda)^{-1} A(\lambda) \right] K \\
&= K - KK^{-1}K \\
&= K - K \\
&= \mathbf{0}.
\end{align*}
Hence, the posterior covariance $K_{N}(\lambda) = \mathbf{0}$. Therefore, the posterior distribution collapses to the trivial solution $\mathbf{u}=\mathbf{0}$.

\paragraph{$\lambda$ is an eigenvalue of $L$:}
We first write the posterior random variable into the form of a linear transformation of the standard Gaussian random variable. Let \(K = PP^\top\) be the Cholesky decomposition of \(K\). Then for the predictive covariance, we have
    \begin{equation}
    \begin{aligned}
        &K - KA^\top(A(\lambda)KA(\lambda)^\top)^{+}(A(\lambda)K)  \\
        &= PP^\top - PP^\top A(\lambda)^\top(A(\lambda) PP^\top A(\lambda)^\top)^{+}(A(\lambda)PP^\top)\\
        & = P\left( I - (A(\lambda)P)^\top(A(\lambda) P (A(\lambda)P)^\top)^{+}A(\lambda)P\right)P^\top\\
        & = P\left( I - ((A(\lambda)P)^{+}A(\lambda)P)^{\top}(A(\lambda)P)^{+}A(\lambda)P \right)P^\top \\
        (\text{Let } Q = A(\lambda)P)\quad& = P\left(I - (Q^{+}Q)^\top(Q^{+}Q)\right)P^\top\\
        & = P(I - Q^{+}Q)P^\top\\
        & = P(I - Q^{+}Q)(I - Q^{+}Q)^\top P^\top
    \end{aligned}
\end{equation}
Note that \(Q^{+}Q\) and \(I - Q^{+}Q\) are orthogonal projection operators, which are symmetric and idempotent. Let \(B = P(I - Q^{+}Q)\), so \(BB^\top = K - KA(\lambda)^\top(A(\lambda)KA(\lambda)^\top)^{+}A(\lambda)K\). Therefore, a sample $\bu^*$ of $\mathcal{N}(0, K_N(\lambda))$ can be written into $\bu^* = B\xi$, where $\xi \sim \mathcal{N}(\mathbf{0},I)$.

We have 
\begin{equation}
    \begin{aligned}
        A(\lambda)\bu^* &= A(\lambda)B\xi \\
        &= A(\lambda)P(I - Q^{+}Q)\xi \\
        &= A(\lambda)P(I - (A(\lambda)P)^{+}(A(\lambda)P))\xi\\
        &= (A(\lambda)P - A(\lambda)P(A(\lambda)P)^{+}(A(\lambda)P))\xi\\
        & = 0
    \end{aligned}
\end{equation}
Hence, any vector $\bu^*$ sampled from the posterior distribution $\mathcal{N}(0,K_{N}(\lambda))$ is in $\text{Null}(A(\lambda))$, which is also the eigenspace of $L$ corresponding to the eigenvalue $\lambda$.

\section{Computational Complexity and Future Optimisations}
\label{appendix:computation}

Here is a formal algorithm that outlines the procedure.
\begin{algorithm}[H]
\caption{Eigenvalue Scanning via PI-GPR Posterior Covariance}
\label{alg:eigen_scan_v2}
\begin{algorithmic}
\Require Differential operator $\calL$, prior covariance function $k(\mathbf{x}, \mathbf{x}')$, observation points $X_N = \{\bx_i\}_{i=1}^{N}$, test points $X_t = \{\bx_i\}_{i=1}^{N_t}$, scanning range $[\lambda_{\min}, \lambda_{\max}]$, and number of steps $N_\lambda$. 
\vspace{1ex}
\State Initialize results list $\mathcal{S} = \emptyset$.
\State Pre-compute prior covariance at test points: $\mathbf{K}_{tt} = k(X_t, X_t)$.
\State Initialize counter $k=1$.
\vspace{1ex}

\While{$k \le N_\lambda$} 
    \State 1. Set the spectral parameter:
    $$\lambda_k = \lambda_{\min} + (k-1)\frac{\lambda_{\max}-\lambda_{\min}}{N_\lambda-1}.$$ \vspace{-1.5ex}
    \State 2. Define the operator $\calL_{\lambda_k} = \calL - \lambda_k I$ and construct the covariance matrices:
    \begin{align*}
     \mathbf{K}_{NN}(\lambda_k) &= \calL_{\lambda_k} \calL_{\lambda_k}' k(X_N, X_N) \\
     \mathbf{K}_{tN}(\lambda_k) &= \calL_{\lambda_k}' k(X_t, X_N)
    \end{align*} \vspace{-1.5ex}
    \State 2. Compute the posterior covariance matrix at the test points:
    $$\mathbf{K}_{N}(\lambda_k) = \mathbf{K}_{tt} - \mathbf{K}_{tN}(\lambda_k) \left(\mathbf{K}_{NN}(\lambda_k)\right)^{+} \mathbf{K}_{tN}(\lambda_k)^\top.$$ \vspace{-1.5ex}
    \State 3. Compute the trace metric and store the result:
    $$J(\lambda_k) = \text{Tr}(\mathbf{K}_{N}(\lambda_k)).$$
    \State \hspace{1.5em} Append $(\lambda_k, J(\lambda_k))$ to $\mathcal{S}$. \vspace{0.5ex}
    \State 4. Increment counter: $k \leftarrow k+1$.
\EndWhile

\Return A set of spectral data points $\mathcal{S} = \{(\lambda_k, J(\lambda_k))\}_{k=1}^{N_\lambda}$.
\end{algorithmic}
\end{algorithm}

In Algorithm \ref{alg:eigen_scan_v2}, we present a naive implementation as a realisation of our eigenvalue criterion. We emphasise that the primary contribution of this work is the introduction of a novel framework for exploring eigenvalue problems through the lens of GP regression. The computational cost of this baseline algorithm is dominated by two components, the spectral scanning over the domain of $\lambda$ ($N_\lambda$ iterations) and the inversion of the dense $N \times N$ covariance matrix ${K}_{NN}(\lambda)$. Consequently, the total complexity scales as $\mathcal{O}(N_\lambda N^3)$. However, the flexibility of the GP framework offers substantial space for future optimisation to reduce this cost. To transition this framework toward large-scale or high-dimensional problems, the following strategies may be considered:
\begin{itemize}
    \item Reducing the Spectral Search Cost ($N_\lambda$): 
    Instead of a brute-force grid search, the exploration of $\lambda$ can be accelerated using certain strategies. For example, since the trace metric $J(\lambda)$ is differentiable, the eigenvalue identification can be formulated as a gradient-based optimisation problem. The Grid Search has a complexity of $\mathcal{O}(1/\epsilon)$ (where $\epsilon$ is the desired precision), whereas Gradient-based methods (like BFGS or Newton-CG) typically exhibit superlinear or quadratic convergence, scaling as $\mathcal{O}(\log(1/\epsilon))$ or $\mathcal{O}(\log \log(1/\epsilon))$ \cite{nocedal2006numerical}. Alternatively, one could leverage the smoothness of the posterior statistics to build a \textit{GP emulator} that treats $\lambda$ as an input dimension, predicting the entire spectrum $J(\lambda)$ from a small number of expensive evaluations \cite{bai2024gaussian}.
    \item Reducing Data Dimensionality ($N$):
    In this paper, we only introduced two types of covariance functions. However, there are a variety of properties that can be embedded into the covariance functions\cite{ginsbourger2016degeneracy, reisert2007learning}. Their pre-embedded properties could allow the GP to achieve the same accuracy with much fewer collocation points \cite{glielmo2017accurate}. Another example is the Hilbert space GP framework proposed by \cite{solin2020hilbert}, which approximates the GP prior using a truncated basis expansion of the Laplacian eigenfunctions specific to the domain geometry. By projecting the process onto a fixed set of $m$ basis functions (where $m \ll N$), the method transforms the non-parametric GP into a parametric linear model, which reduces the scaling from $\mathcal{O}(N^3)$ to $\mathcal{O}(N m^2)$. We can also develop operator-specific kernels derived from fundamental solutions or Green’s functions \cite{alvarez2013linear}. By embedding the analytical properties of the underlying physical operator into the GP covariance structure, the resulting Kernels could potentially sharpen the trace peaks $J(\lambda)$, thereby increasing the resolution of the eigenvalue detection while requiring fewer spatial collocation points.
    \item Accelerating Linear Algebra (The Cubic Term): 
    To mitigate the $\mathcal{O}(N^3)$ cost of inversion, we can employ Sparse GP approximations. Methods based on inducing points (e.g., FITC, VFE) project the process onto a lower-dimensional set of $M$ variables ($M \ll N$), reducing the complexity to $\mathcal{O}(N M^2)$. For structured grids, methods such as Structured Kernel Interpolation (SKI) can further reduce operations to $\mathcal{O}(N \log N)$ via fast Fourier transforms.
\end{itemize}

\end{document}